\documentclass{article}
\usepackage{caption}

\usepackage[textsize=tiny]{todonotes}

\usepackage{microtype}
\usepackage{graphicx}
\usepackage{subfigure}
\usepackage{booktabs} 

\usepackage{algorithm} 
\usepackage{algpseudocode} 
\usepackage{hyperref}

\usepackage{amsmath,amsfonts,bm}

\def\eqref#1{equation~\ref{#1}}

\def\1{\bm{1}}

\DeclareMathAlphabet{\mathsfit}{\encodingdefault}{\sfdefault}{m}{sl}
\SetMathAlphabet{\mathsfit}{bold}{\encodingdefault}{\sfdefault}{bx}{n}

\def\gA{{\mathcal{A}}}

\def\gL{{\mathcal{L}}}
\def\gM{{\mathcal{M}}}

\def\gS{{\mathcal{S}}}

\usepackage[accepted]{icml2024}
\usepackage{amsmath}
\usepackage{amssymb}
\usepackage{mathtools}
\usepackage{amsthm}
\usepackage{enumerate}   

\usepackage[capitalize,noabbrev]{cleveref}
\usepackage[normalem]{ulem}
\usepackage{booktabs}
\theoremstyle{plain}
\newtheorem{theorem}{Theorem}[section]

\theoremstyle{definition}
\newtheorem{definition}[theorem]{Definition}

\theoremstyle{remark}

\usepackage{colortbl}
\usepackage{booktabs}
\usepackage{hhline}
\usepackage{color}
\usepackage{multirow}
\usepackage{diagbox}
\usepackage{arydshln}

\usepackage{ragged2e}
\usepackage{makecell}
\usepackage{tcolorbox}
\usepackage{CJKutf8}
\usepackage[utf8]{inputenc}

\usepackage{booktabs}

\usepackage[toc,page,header]{appendix}
\usepackage{minitoc}

\usepackage{listings}

\newcommand{\FT}{\text{FT}}

\icmltitlerunning{Selecting Large Language Model to Fine-tune via Rectified Scaling Law}

\begin{document}

\twocolumn[
\icmltitle{Selecting Large Language Model to Fine-tune via Rectified Scaling Law}



\icmlsetsymbol{equal}{*}

\begin{icmlauthorlist}
\icmlauthor{Haowei Lin}{equal,pkuaia,pku}
\icmlauthor{Baizhou Huang}{equal,pku}
\icmlauthor{Haotian Ye}{equal,stf}
\icmlauthor{Qinyu Chen}{pku}
\icmlauthor{Zihao Wang}{pkuaia,pku}\\
\icmlauthor{Sujian Li}{pku}
\icmlauthor{Jianzhu Ma}{thu}
\icmlauthor{Xiaojun Wan}{pku}
\icmlauthor{James Zou}{stf}
\icmlauthor{Yitao Liang}{pkuaia,pku}
\end{icmlauthorlist}

\icmlaffiliation{pkuaia}{Institute for Artificial Intelligence, Peking University}
\icmlaffiliation{pku}{Peking University}
\icmlaffiliation{stf}{Stanford University}
\icmlaffiliation{thu}{Tsinghua University}
\icmlcorrespondingauthor{Yitao Liang}{yitaol@pku.edu.cn}

\icmlkeywords{Large Language Model, Model Selection, Scaling Law}

\vskip 0.3in
]



\printAffiliationsAndNotice{\icmlEqualContribution} 

\begin{abstract}
The ever-growing ecosystem of LLMs has posed a challenge in selecting the most appropriate pre-trained model to fine-tune amidst a sea of options. 
Given constrained resources, fine-tuning all models and making selections afterward is unrealistic. 
In this work, we formulate this resource-constrained selection task into predicting fine-tuning performance and illustrate its natural connection with Scaling Law. 
Unlike pre-training, we find that the fine-tuning scaling curve includes not just the well-known ``power phase'' but also the previously unobserved ``pre-power phase''. We also explain why existing Scaling Law fails to capture this phase transition phenomenon both theoretically and empirically.
To address this, we introduce the concept of ``pre-learned data size'' into our Rectified Scaling Law, which overcomes theoretical limitations and fits experimental results much better. 
By leveraging our law, we propose a novel LLM selection algorithm that selects the near-optimal model with hundreds of times less resource consumption, while other methods may provide negatively correlated selection. The project page is available at \url{rectified-scaling-law.github.io}.

\end{abstract}

\section{Introduction}
\label{sec:intro}

Recent years have witnessed the unprecedented development of large language models (LLMs)~\citep{touvron2023llama,Achiam2023GPT4TR}, as well as the benefits they bring to numerous downstream tasks~\citep{liu2019roberta,devlin2018bert}. Among all progresses, one important technique is \textit{fine-tuning}, which re-trains a pre-trained model on specific datasets to convert the model into a task-specific expert~\citep{Ke2023AdaptingAL,Ke2023ContinualPO}. It has been widely demonstrated that fine-tuning can substantially improve the performance of downstream applications~\citep{raffel2020exploring,alt2019fine}. The common workflow of fine-tuning a LLM starts with selecting an appropriate pre-trained model. Thanks to the ever-growing ecosystem of LLMs like HuggingFace, we are able to choose from countless models for specific downstream task fine-tuning. 



\begin{figure*}[!t]
    \centering
    \includegraphics[width=\textwidth]{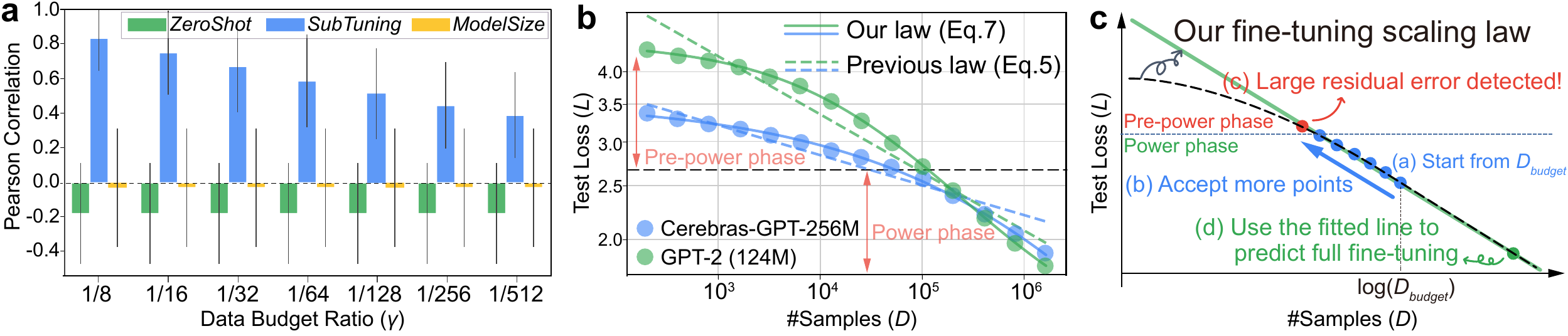}
    \vspace{-2em}
    \caption{(\textbf{a}) The Pearson correlation between the true full-fine-tuning performance and the predicted performance of three intuitive methods, given different resource constraints denoted by $\gamma$. These baseline methods cannot predict performance well especially under demanding constraints (small $\gamma$), and could even provide negatively correlated predictions.
    (\textbf{b}) The phase transition phenomenon observed in the scaling of fine-tuning loss $L$ with training sample size $D$. In addition to the widely studied power phase where $(L,D)$ are linearly correlated under the log-log scale, we discover the pre-power phase when $D$ is small. Previous laws fail to fit both phases, while our proposed law fits quite well.
    (\textbf{c}) Our LLM selection algorithm that extrapolates full-fine-tuning performance based on the new law. 
    }
    \label{fig:intro}
    \vspace{-5pt}
\end{figure*}

However, the explosion of open-sourced models also poses a ``mixed blessing'': how can we select the model with optimal performance after fine-tuning? Given various resource constraints on time, computation and storage~\citep{hoffmann2022training}, it is unrealistic to fine-tune all candidates and make selections afterward. It is also unstable and unpredictable to rely on empirical human impressions to select LLM for a new task, such as selecting the largest one, the most well-known one, or even the one with the highest zero-shot performance on targeted tasks~\citep{brown2020language}. 
In addition, most existing model selection methods~\citep{Vu2020ExploringAP,Dwivedi2020DualityDS} fail to solve LLM fine-tuning tasks because they were designed for classification and regression tasks, which is incompatible with generative LLMs~\citep{bai2023determine}.
This brings us to the problem of LLM selection for fine-tuning from a unified perspective, especially in a resource-constrained manner. 

To better address this challenge, we formulate \textit{LLM Selection in the context of fine-tuning} for the first time. Our framework models the challenge as a resource-constrained task to predict the \textit{full-fine-tuning} performance of a model, $i.e.$, the performance after fine-tuning on the entire downstream task dataset. 
By measuring the error between the predicted and the true full-fine-tuning performance, we further show that intuitive selection methods based on model size, zero-shot performance, or fine-tuned performance on a small subset, all fail to give a good full-fine-tuning performance prediction (\cref{fig:intro}(a)). The correlation between their prediction and the ground-truth performance is surprisingly low.

We point out that the challenge in predicting full-fine-tuning performance with limited resources naturally draws parallels to the study of \emph{LLM Scaling Law}~\citep{kaplan2020scaling}, which has been successfully applied to predict the LLM pre-training performance with at most $10,000\times$ less compute~\citep{Achiam2023GPT4TR}. Similarly, \emph{can we leverage Scaling Law to efficiently and accurately predict the performance of fine-tuning as well?}

In this paper, we conduct thorough experiments on scaling behavior in fine-tuning using $30$ models with sizes varying from $100$ million to $7$ billion. As shown in \cref{fig:intro}(b), we find a previously unobserved phase transition pattern called ``\textit{pre-power phase}'' on the low-data regimes where the slope gradually decreases before the widely studied ``power phase'' where the test loss and number of samples $D$ is roughly linearly correlated. The transition is crucial for fine-tuning, as typical fine-tuning datasets can vary from hundreds to millions of samples, covering both phases.
We theoretically explain this phenomenon via the concept of \textit{pre-learned data size}, which represents the equivalent amount of downstream task samples that the model has pre-learned from the pre-training corpus. Inspired by this, we establish {Rectified Scaling Law of LLM fine-tuning} ($a.k.a.$ ``Fine-tuning Scaling Law'') by incorporating this concept (\cref{eq.law}), which fits all experimental results much better than all existing laws, aligning with our theoretical judgments.  

Based on the Rectified Scaling Law of LLM fine-tuning, we design a novel LLM selection algorithm called ``Accept then Stop'' (\textit{AtS}, \cref{fig:intro}(c)). Starting from the maximum allowed constraints, it keeps accepting fine-tuning results on a series of size-decreasing subsets, stops once it distinguishes the transition pattern, and uses all accepted results to linearly extrapolate the full-fine-tuning performance. The designed algorithm demonstrates outstanding LLM selection performance under extensive experimental settings, and selects the near-optimal model with hundreds of times less resource consumption, under which other approaches can provide negatively correlated selection results. Extensive ablation experiments also prove its robustness and stability.

In summary, we first formulate LLM selection framework with great compatibility, and draw its connection with the study of Scaling Law for model fine-tuning (\cref{sec:prelim}). We demonstrate why previous laws fail to fit fine-tuning performance both theoretically and empirically, and establish a new Scaling Law that fits much better (\cref{sec.ftlaw}). We propose a novel LLM selection algorithm based on the established law that significantly outperforms all other baselines under extensive experimental settings (\cref{sec.llmselection}). Together, our work makes a first step towards LLM selection for fine-tuning, and towards better understanding of Scaling Law in practical downstream applications.

\section{LLM Selection Framework for Fine-tuning}
\label{sec:prelim}
\subsection{Problem Formulation}

\label{subsec.preliminary_llmselection}
Throughout the paper, we consider the standard \textit{supervised fine-tuning} \citep{dai2015semisupervised,devlin2018bert} paradigm in full parameter space of \textit{auto-regressive models} \citep{graves2014generating} that sequentially predicts each token in target $\boldsymbol{y}$ based on input $\boldsymbol{x}$. For a pre-trained model $M$ and a dataset $\mathcal S$, we use $\text{FT}(M;\mathcal S)$ to denote the fine-tuned model on dataset $\mathcal S$ from $M$\footnote{The fine-tuning process is regarded as a black box in our paper as our focus is not on ``how to fine-tune a model''.} 
We formulate model selection task in the context of fine-tuning as follows.


\begin{definition}[LLM Selection for Fine-tuning]\label{def:framework}
    Given a set of pre-trained LLMs $\mathcal M=\{M_i\}_{i=1}^m$ with $m$ models, a fine-tuning sub-dataset $\gS_{sub}$ sampled from the complete dataset $\gS$, i.e., $\gS_{sub} \subset \gS \sim \mathbb D, |\gS_{sub}| = \gamma |\gS|$ where $\gamma \in(0,1]$ is the data budget ratio, the goal of an LLM selection algorithm $\mathcal A: (M;\gS_{sub}) \mapsto \mathbb R$ is to score each model $M \in \mathcal M$ with access to $\gS_{sub}$, such that the score reflects the loss over distribution $\mathbb D$ after fine-tuning $M$ on $\mathcal S$, i.e., we hope that
        \begin{align}
        \gL(\FT(\hat M(\gS_{sub}); \gS)) &= \min_{M \in \mathcal M} \gL (\FT(M;\gS)),\\
        \text{where}\quad \hat M(\gS_{sub}) &\triangleq \mathop{\arg\min}_{M\in \mathcal M} \gA(M, \gS_{sub}).
        \label{eq.llm_selection}
        \end{align}
\end{definition}

Here $\gL(M)$ is the expectation of the average of cross-entropy loss of model $M$ on sample $(\boldsymbol{x},\boldsymbol{y})$ over the target token sequence $\boldsymbol{y}$, i.e.
\begin{equation}
    \gL(M) = \mathbb E_{(\boldsymbol x, \boldsymbol{y})\sim \mathbb D} -    \frac{1}{|\boldsymbol{y}|}\sum_{j=1}^{|\boldsymbol{y}|}\log(P(y_j|\{y_i\}_{i=1}^{j-1}, \boldsymbol{x})).
    \label{eq.loss}
\end{equation}
\cref{def:framework} introduces the subset $\gS_{sub}$ to model various types of constraints, for instance from efficiency consideration, in which people want to find the best model without fine-tuning on the entire training set but on a much smaller set to reduce resource consumption. Under the same model set $\gM$, the difficulty of LLM selection is captured by the data budget ratio $\gamma$, where smaller $\gamma$ means we need to predict the performance of $\FT(M,\gS)$ with access to a smaller set $\gS_{sub}$. 
Our framework can also simulate the constraints on model families or GPU memory via the control of $\gM$, making it compatible with practical selection problems with different requirements.

For the sake of the consistency of performance estimation in our study, we safely hold out a validation set and always use the average loss over this set as the estimation of $\gL(M)$ for models fine-tuned on different $\gS_{sub}$. In practice, this set could be obtained by holding out a subset from the training set, as we assume that both the training set and test set are sampled from $\mathbb D$ in an \textit{i.i.d.} way.


\subsection{Connecting to Scaling Law}
Predicting $\gL(\FT(M, S))$ using a subset $\gS_{sub}$ is closely related to understanding the scaling behavior in the \emph{fine-tuning} stage.
Indeed, the Scaling Law in the \emph{pre-training} stage has been widely studied \citep{henighan2020scaling,kaplan2020scaling,bahri2021explaining}, and it is commonly believed to have the form below.

\begin{definition}[Power-law in \cite{kaplan2020scaling}]\label{def:law}
    The scaling loss $\hat \gL(\cdot,\cdot)$ is a function of model size $N$ and training set size $D$, i.e., 
    \begin{equation}
        \hat \gL(N,D) = \Big(\frac{A}{N^{\alpha_N}} + \frac{B}{D^\beta}\Big)^\alpha.
        \label{eq.power-law}
    \end{equation}
\end{definition}

\begin{figure}[t]
    \centering
    \includegraphics[width=\linewidth]{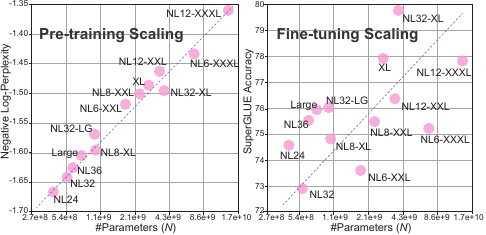}
    \vspace{-2em}
    \caption{The difference of scaling behavior in pre-training and fine-tuning. While in pre-training the performance scales with model sizes independent from model shapes, in fine-tuning the performance does not. The figure is drawn based on Figure 1 in~\citet{tay2021scale}.}
    \label{fig:chaos}
    \vspace{-5pt}
\end{figure}

Here $\{A, B, \alpha, \alpha_N, \beta\}$ are universal parameters to be fitted, and we always use $\hat \gL$ to indicate that this is an estimated function of true losses. While it is universally observed in many tasks and domains when training models from scratch~\citep{ghorbani2021scaling, alabdulmohsin2022revisiting, fernandes2023scaling}, \citet{tay2021scale} finds that the scaling behavior may differ in the fine-tuning phase. As shown in \cref{fig:chaos}, fine-tuning loss is dependent on models not only through their sizes $N$, but also through other inductive bias like model architecture, the number of layers, attention heads, hidden dimensions and so forth.

This observation makes it highly non-trivial to select a model using existing Scaling Law. For instance,~\cref{eq.power-law} implies that models with more parameters work better under the same $\gS$, which is contradictory to \cref{fig:chaos}. 
Fortunately, as our goal is to predict performance for each model, a marginal version of the Scaling Law when the model is fixed is sufficient. 
In this case, the complexity of model architectures can be removed, and the law proposed in \cite{kaplan2020scaling,hernandez2021scaling,tay2021scale} share the following unified form:
\begin{align}\label{eq:prev_law_fix}
    \hat \gL(D) = \big(\frac{B}{D^\beta} + E\big)^{\alpha}.
\end{align}
Here $D$ is the number of training data, and $B, E,\alpha, \beta$ are model/task-dependent parameters. All parameters in \cref{def:law} are non-negative, and so as in \cref{eq:prev_law_fix}. 


\section{Analysis on Fine-tuning Scaling Law}
\label{sec.ftlaw}


In this section, we fine-tune $30$ LLMs on three datasets with a sufficiently wide range of dataset size, and illustrate the existence of the ``phase transition'' pattern during scaling fine-tuning. We demonstrate both theoretically and empirically why \cref{eq:prev_law_fix} fail to fit the results.
Based on our theoretical analysis, we introduce the concept of \textit{pre-learned data size} and establish a well-fitted Scaling Law by incorporating the pre-learned data size into existing laws. 


\subsection{Setup}

We first introduce the experimental settings of models, datasets, optimization, and evaluations. 
These settings are shared across the study of Scaling Law and LLM selection.

\paragraph{LLM Set.} To ensure the comprehensiveness of our study, we choose a wide range of open-sourced LLMs released by different organizations \emph{in the wild}, with various architectures, pre-trained corpus, training strategies, and model sizes. In total, $30$ models with the number of parameters ranging from $100$ million to $7$ billion are selected to form the model set $\mathcal M$. We include both encoder-decoder models such as T5~\citep{raffel2020exploring} and decoder-only models such as GPT2~\citep{Radford2019LanguageMA}. We also include some multilingual models~\citep{xue2021mt5}, MoE-based models~\citep{fedus2022switch}, and instruction-tuned models~\citep{wu2024laminilm} for diversity.
For clarity, we select $6$ representative models for illustrations throughout the main paper, including GPT2, LaMini-GPT (the instruction-tuned version of GPT2), Cerebras-GPT (three different versions for comparison) and mT5 (a multilingual encoder-decoder model). 
Results of complete model set are presented in~\Cref{app.llm}.

\paragraph{Fine-tuning Datasets.}
We consider machine translation (WMT19 English-Chinese (En-Zh)~\citep{kocmi2022findings}), paragraph summarization (Gigaword~\citep{Rush_2015}), and multi-task instruction tuning (FLAN~\citep{wei2021finetuned}) as the downstream fine-tuning tasks. These tasks are representative and well-established in NLP with rich amount of data, allowing us to study the scaling behavior under a wide range of dataset size. Details of the processing of each dataset are presented in~\Cref{app.datasets}.

\paragraph{Dataset Size.} To study the scaling behavior extensively, for each dataset $\gS$, we randomly select subsets with $D$ samples where $D\in\{200, 400, 800, \cdots, 1638400\}$ which cover a wide range of data scales in practical scenarios. We fine-tune models on each subset and test them on a held-out test set with  samples to ensure the estimated performance is unbiased. For each setting, we fine-tune the model three times to remove the randomness of subset sampling.

\paragraph{Optimization.} We adopt the standard fine-tuning using AdamW~\citep{loshchilov2017decoupled} optimizer and cosine learning rate scheduler~\citep{Loshchilov2016SGDRSG}. We optimize each model under different initial learning rates and batch sizes via hyper-parameter search. This ensures that test losses are optimal under current settings. More details of fine-tuning are presented in~\Cref{app.ftsetting}.

\begin{figure*}[th]
    \centering
    \includegraphics[width=\textwidth]{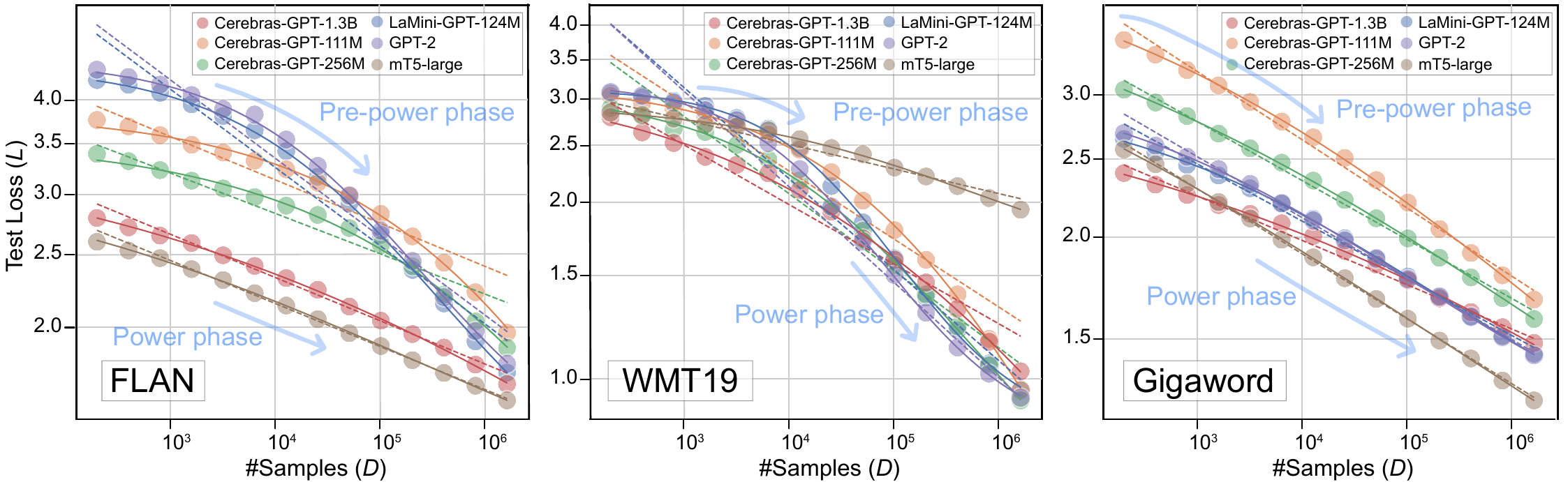}
    \vspace{-2em}
    \caption{The phase transition from pre-power phase to power phase, and the fitness of different Scaling Laws. The x and y axes are fine-tuning dataset size $D$ and test loss $L$ in log scale. Each subfigure corresponds to a dataset.  Solid lines are the fitting results of our law (Eq. \ref{eq.law}), and dash lines are the fitting results of vanilla law (Eq. \ref{eq:prev_law_fix}). The full model results are in~\Cref{app:fine-tune-results}.}
    \label{fig:sec3-main-law}
\end{figure*}

\subsection{Phase Transition with Dataset Size}

We plot the test loss for 6 representative models when fine-tuned on subsets of different sizes in \cref{fig:sec3-main-law}. We observe a ``phase transition'' pattern in scaling behaviors: when the loss is relatively large, the curve lies in ``pre-power phase with the slope of the curve slowly decreases; as the training set size $D$ increases, the loss decreases and the curve enters the ``power phase'' where it is almost linear, similar to the observed curves in the pre-training stage. For different datasets, depending on their difficulty, the size of data each model requires to transit into the second phase is different. 

The pre-power phase has been barely observed before, mainly due to the focus on large data regimes. Indeed, for scaling behavior in pre-training or language-to-code transfer \citep{hernandez2021scaling} in which the minimum sample size is $\sim10^5$, models have already entered the power phase and the pre-power phase becomes invisible. However, many fine-tuning tasks may fall into a relatively low-data regime, making the analysis of the behavior of the pre-power phase inevitable. Below we show that \cref{eq:prev_law_fix} does not take this phase into consideration.


\begin{theorem}\label{thm:derivative}
    For any positive parameters $B, E, \alpha, \beta$, consider the log-log form of function $\hat \gL(\cdot)$ in \cref{eq:prev_law_fix}: 
    \begin{align}
        f(x) = \log(\hat \gL(\exp(x)) = \alpha \log \big(\frac B {\exp(\beta x)}+E\big),
    \end{align}
    then we have that the derivative $f'$ is negative and non-decreasing.
\end{theorem}

\cref{thm:derivative} establishes a crucial property that the slope of $f'$ cannot decrease, contradictory to the co-existence of pre-power and power phase, since slopes \textit{decrease} initially and remain roughly unchanged afterward. Indeed, as demonstrated in \cref{fig:sec3-main-law}, it fits poorly with experimental results (dash lines)\footnote{The parameters are fitted using a standard python optimization package, and please refer to Appendix \ref{app.optimization} for more details.}, manifesting by the deviation of the predicted loss and actual loss in the pre-power phase.

\subsection{Our Scaling Law with Pre-learned Data}
To better understand the underlying mechanism of the phase transition phenomenon, we start with the essential difference between pre-training and fine-tuning. Unlike pre-training where we train a model \textit{from scratch}, fine-tuning starts from a model that has been trained on a large corpus. Consequently, pre-training should have provided models with some amount of information relevant to downstream context~\citep{hernandez2021scaling}.

To capture this concept, we introduce the term \textit{pre-learned data size} (represented by $D_l$) that indicates how much amount of downstream data a model has learned from pre-training. This term could be influenced by multiple factors like the expressivity of models, the pre-training corpus size, as well as the difficulty of this downstream task. Intuitively, $D_l$ can be integrated with the scaling term $D^\beta$, which represents the amount of information that fine-tuning on $D$ samples can provide the model with. We propose the following improved Scaling Law by incorporating this term, with an identical amount of parameters to be fitted.

\begin{definition}[Rectified Scaling Law]\label{def:our_law}
    We define the Scaling Law with dataset size $D$ for fine-tuning as 
    \begin{align}
        \hat \gL(D) = \frac{B}{D_l+D^\beta} + E,
        \label{eq.law}
    \end{align}
    where $D_l$ is the pre-learned data size, $\beta$ is the power to $D$ denoting the learning difficulty, $B$ adjusts the initial test loss, and $E$ denotes the optimal loss of the model given an infinite amount of data.\footnote{The parameter $\alpha$ is unnecessary for our law to fit well, and we remove it for the sake of simplicity. All parameters are model-specific. We leave the incorporation of model information into \cref{def:our_law} as future explorations.}
\end{definition}

This modification of $D_l$ essentially improves the mathematical property of \cref{def:our_law} as the derivative is no longer monotonous. As proved in \cref{thm:second}, the first-order derivative decreases before $x_0$ (corresponding to the pre-power phase) and slightly increases afterward (corresponding to the power phase). In other words, the introduction of $D_l$ is not only conceptually reasonable, but it also elegantly unifies the co-existence of both phases into one law. 

\begin{theorem}\label{thm:second}
    For any positive parameters $B,E,D_l,\beta$, consider the log-log form of function $\hat \gL(\cdot)$ in \cref{eq.law}:
    \begin{align}
        f(x) = \log(\hat \gL(\exp(x))  = \log\big(\frac B {D_l+\exp(\beta x)} + E\big),
    \end{align}
    then the second-order derivative $f''$ is negative for $x\in (0, x_0)$ and positive for $x \in (x_0,+\infty)$, where we have $x_0 = \frac{\log(D_l^2 + BD_l/E)}{2\beta}$.  
\end{theorem}
\begin{figure}[t]
    \centering
    \includegraphics[width=0.9\linewidth]{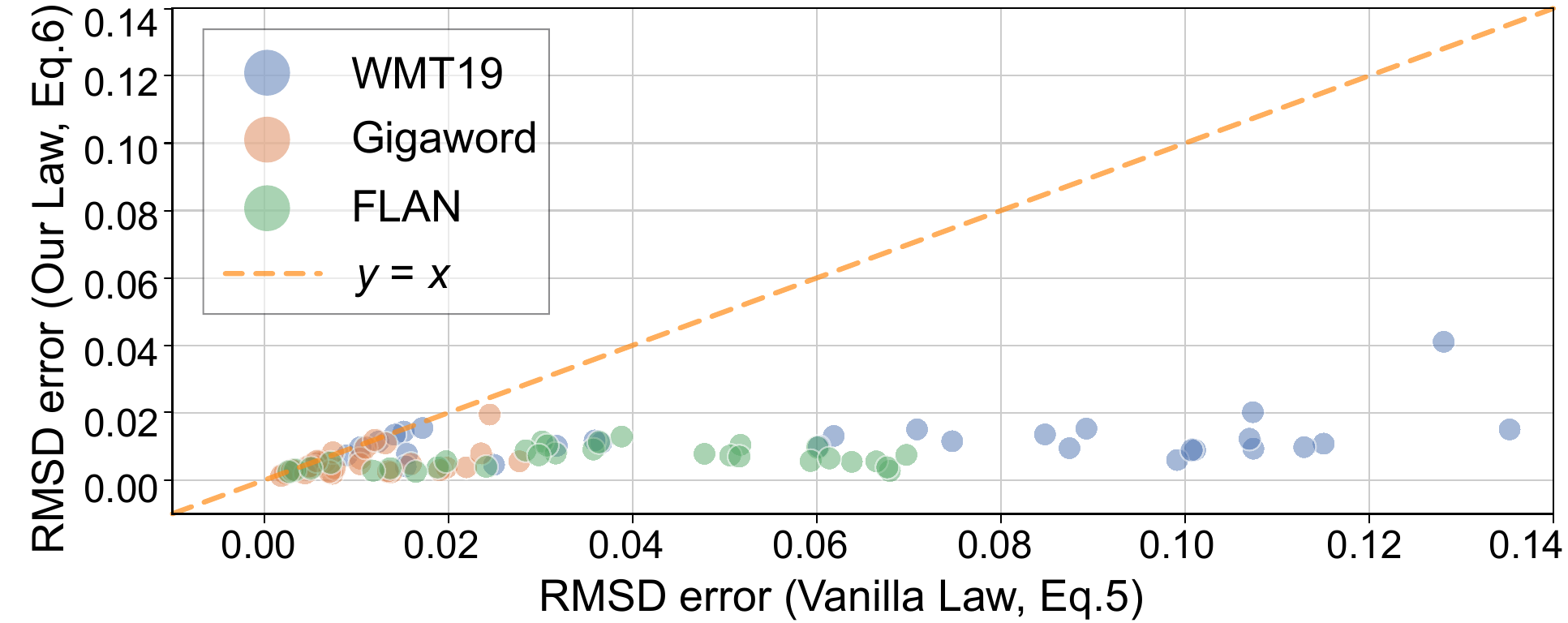}
    \vspace{-1em}
    \caption{Root-mean-square deviation (RMSD) of our law (\cref{eq.law}) and vanilla law (\cref{eq:prev_law_fix}) when fitting fine-tuning test loss versus dataset size in log scale. Under same setting, our law achieves much lower RMSD error.}
    \label{fig:error_comparison}
    \vspace{-5pt}
\end{figure}

We quantified the fitting error of both laws on all models and datasets using root-mean-square deviation (RMSD) in \cref{fig:error_comparison}. On average, each law is required to fit fifteen size-loss pairs. The error of \cref{eq:prev_law_fix} is unavoidably large (with an average RMSD of $0.036$). As it can only fit the power phase, a more difficult task results in a later occurrence of phase transition, contributing to a larger fitting error. On the contrary, our law \cref{eq.law} has a consistently small RMSD error, with an average RMSD of $0.007$. Since both laws have four parameters to fit, it demonstrates that our law captures the intrinsic scaling behavior more accurately.






\begin{table*}[t]

    \renewcommand{\arraystretch}{1.1}
    \centering
    \caption{Model selection results (\textbf{PearCorr}, \textbf{RelAcc}) of four methods on three datasets (FLAN, WMT19, Gigaword) in percentage. The best result within the same dataset and budget ratio is in \textbf{bold} font, and the second best result is \underline{underlined}.
}
    \Huge
    \resizebox{\linewidth}{!}{\begin{tabular}{c|c|cccc|cccc|cccc}
\hline

\multicolumn{2}{c|}{} & \multicolumn{4}{c}{{\textbf{FLAN}}} & \multicolumn{4}{c}{\textbf{WMT19}} & \multicolumn{4}{c}{\textbf{Gigaword}}\\
Metric & ~~~~Ratio~~~~ &
~~~~\textit{AtS}~~~~ & \textit{ZeroShot} & \textit{SubTuning} & \textit{ModelSize} & ~~~~\textit{AtS}~~~~ & \textit{ZeroShot} & \textit{SubTuning} & \textit{ModelSize} & ~~~~\textit{AtS}~~~~ & \textit{ZeroShot} & \textit{SubTuning} & \textit{ModelSize} \\
\hline
\multirow{8}{*}{\textbf{PearCorr} (\%)} &1/8 & \textbf{90.9} & -10.7 & \underline{60.9} & -20.9 & \textbf{98.9} & 7.1 & \underline{93.5} & 36.0 & \textbf{98.9} & -49.2 & \underline{93.2} & -24.4 \\
~ &1/16 & \textbf{73.1} & -10.7 & \underline{46.5} & -20.9 & \textbf{97.1} & 7.1 & \underline{87.1} & 36.0 & \textbf{97.6} & -49.2 & \underline{89.3} & -24.4 \\
~ &1/32 & \textbf{65.5} & -10.7 & \underline{36.4} & -20.9 & \textbf{97.7} & 7.1 & \underline{77.7} & 36.0 & \textbf{96.9} & -49.2 & \underline{85.4} & -24.4 \\
~ &1/64 & \textbf{61.1} & -10.7 & \underline{29.0} & -20.9 & \textbf{86.0} & 7.1 & \underline{64.5} & 36.0 & \textbf{92.0} & -49.2 & \underline{80.9} & -24.4 \\
~ &1/128 & \textbf{52.2} & -10.7 & \underline{24.5} & -20.9 & \textbf{78.0} & 7.1 & \underline{51.7} & 36.0 & \textbf{91.1} & -49.2 & \underline{76.2} & -24.4 \\
~ &1/256 & \textbf{50.5} & -10.7 & \underline{20.9} & -20.9 & \textbf{73.4} & 7.1 & \underline{41.6} & 36.0 & \textbf{89.1} & -49.2 & \underline{69.9} & -24.4 \\
~ &1/512 & \textbf{45.6} & -10.7 & \underline{16.4} & -20.9 & \textbf{61.5} & 7.1 & 34.5 & \underline{36.0} & \textbf{91.0} & -49.2 & \underline{64.8} & -24.4 \\
~ & Avg & \textbf{62.7} & -10.7 & \underline{33.5} & -20.9 & \textbf{84.6} & 7.1 & \underline{63.4} & 36.0 & \textbf{93.8} & -49.2 & \underline{79.9} & -24.4 \\
\hline

\multirow{8}{*}{\textbf{RelAcc} (\%)} &1/8 & \textbf{93.6} & 85.3 & \underline{93.2} & 59.6 & \textbf{99.1} & 84.4 & \textbf{99.1} & 22.5 & \textbf{100.0} & 71.3 & \underline{87.6} & 71.3 \\
~  &1/16 & \textbf{93.2} & 85.3 & \textbf{93.2} & 59.6 & \textbf{99.1} & 84.4 & \textbf{99.1} & 22.5 & \textbf{91.4} & 71.3 & \underline{87.6} & 71.3 \\
~  &1/32 & \textbf{93.2} & 85.3 & \textbf{93.2} & 59.6 & \textbf{99.6} & 84.4 & \underline{99.1} & 22.5 & \textbf{94.3} & 71.3 & \underline{87.6} & 71.3 \\
~  &1/64 & \textbf{93.2} & 85.3 & \textbf{93.2} & 59.6 & \textbf{99.1} & 84.4 & \textbf{99.1} & 22.5 & \textbf{100.0} & 71.3 & 71.3 & 71.3 \\
~  &1/128 & \textbf{85.3} & \textbf{85.3} & 59.6 & 59.6 & \textbf{99.1} & 84.4 & \textbf{99.1} & 22.5 & \textbf{94.3} & 71.3 & 71.3 & 71.3 \\
~  &1/256 & \textbf{93.2} & \underline{85.3} & 59.6 & 59.6 & \textbf{99.1} & 84.4 & \textbf{99.1} & 22.5 & \textbf{94.3} & 71.3 & 71.3 & 71.3 \\
~  &1/512 & \textbf{93.2} & \underline{85.3} & 59.6 & 59.6 & \textbf{99.1} & 84.4 & \textbf{99.1} & 22.5 & \textbf{91.4} & 71.3 & 71.3 & 71.3 \\
~ & Avg & \textbf{92.1} & \underline{85.3} & 76.4 & 59.6 & \textbf{99.2} & 84.4 & \underline{99.1} & 22.5 & \textbf{95.1} & 71.3 & \underline{79.4} & 71.3 \\
\hline

\end{tabular}}

    \label{tab:main_results}
\end{table*}

\section{LLM Selection}
\label{sec.llmselection}

With a fine-grained understanding of the scaling behavior, we turn to the LLM selection task and propose a novel algorithm that leverages the newly established LLM Fine-tuning Scaling Law. This allows us to select near-optimal models with hundreds of times less resource consumption. 


\subsection{Method: from Scaling Law to LLM Selection}

From the view of Scaling Law, the goal of the LLM selection is to predict subsequent curve given points that can be computed via $\gS_{sub}$. 
We capture the essential ``phase transition'' phenomenon and propose the ``Accept then Stop'' (AtS) algorithm that distinguishes samples from two phases and extrapolates the \emph{power phase}, which is approximately linear under the log-log scale. This algorithm turns out to be more robust and accurate than fitting the entire law directly, which can be sensitive when $\gamma$ is small.


We illustrate the process of \textit{AtS} in \cref{alg:ats}. Specifically, it first fine-tunes the model on $\gS_{sub}$ to compute the test loss. It then continuously reduces the dataset size by half, and fine-tunes the model on this smaller subset to get a series of loss-size pairs $P = \{(\tilde D_i, \tilde L_i)\}$. Whenever a new pair is added, \textit{AtS} fits a linear function $f$ with all previous pairs, and computes stop indicator $I_{stop}$ which captures how deviated the new pair is to the linear function, i.e., 
\begin{equation}
    I_{stop}(\tilde D, \tilde L) \triangleq (|\log {\tilde L} - f(\log {\tilde D})|) / \sigma.
\end{equation}
Here $\sigma$ is the standard deviation of the fitting residuals. \textit{AtS} begins with the first $k$ loss-size pairs accepted without constraints. Then, it keeps accepting new pairs until $I_{stop}$ is larger than a threshold $\delta$, which indicates the occurrence of the pre-power phase. We will use all accepted pairs to fit a linear function $f$ and predicts the full fine-tuning test loss as $\exp(f(\log |\gS|))$. We run all experiments with $k=3$ and $\delta=5$, and conduct ablation studies below.

\begin{algorithm}[t]
	\caption{Accept then Stop (\textit{AtS})}
    \textbf{Input}: Training subset $\gS_{sub}$, Model $M$, parameters $k,\delta$. 
	\begin{algorithmic}[1]
        \State Initialize loss-size pair set $P = \{\}$.
        \While{Ture}
        \State Fine-tune $M$ on $\gS_{sub}$ and get its loss $\tilde L$.
        \If{$|P| \geq k$}
        \State Fit a linear regression model $f$ on $P$.
        \State \textbf{break} if $I_s > \delta$.
        \EndIf
        \State Add pair $\{\log|\gS_{sub}|, \log {\tilde L}\}$ to $P$.
        \State Sample new $\gS_{sub}$ with half size from $\gS_{sub}$.
        \EndWhile
	\end{algorithmic} 
    \textbf{Return:} Score of $M$ as negative predicted $\log$-loss on $\gS$, $-f(\log(|\gS|)$.
 \label{alg:ats}
\end{algorithm}

\subsection{Experimental Settings}
We now formally introduce the baseline methods we compare \textit{AtS} with, and the metrics we use to evaluate each method. The settings of LLM set, datasets, and fine-tuning processes are discussed in \cref{sec.ftlaw}.

\textbf{LLM Selection Baselines.} Notice that we use the data budget ratio $\gamma = \frac{|\gS_{sub}|}{D} \in (0,1]$ to represent the difficulty of a selection task. It can also capture how much faster we want the selection algorithm to be when compared with full-fine-tuning. 
We set $\gamma = \{\frac 1 {512}, \frac 1 {256}, \cdots, \frac 1 8\}$. For comparison, we choose three baseline algorithms $\gA$: (1) \emph{ModelSize} uses the logarithm of the number of model parameters $\log(N)$ as the selection score. (2) \emph{ZeroShot} adopts the zero-shot performance as the selection score; (3) \emph{SubTuning} uses the performance of the subset fine-tuned model $\text{FT}(M,\mathcal S_{sub})$ as the selection score. All the performance is tested on a held-out validation set.


\textbf{Evaluation Metrics.} All selection algorithms give a score to each model $M\in\gM$, and we hope that models with higher scores have better performance when fine-tuned on $\gS$. We consider two metrics below: (1) Pearson correlation coefficient (\textbf{PearCorr}) between scores and full-fine-tuning performance, which measures how we can use the predicted score to rank models. (2) Relative Accuracy (\textbf{RelAcc}), which is defined as the performance gap between the selected model and the best model over the gap between the worst model and the best model, i.e., 
\begin{align*}
    \textbf{RelAcc}(\gA) \triangleq \frac{\max \gL(M)-\gL(\mathop{\arg\max} \gA(M, \gS_{sub}))}{\max \gL(M)- \min \gL(M)}.
\end{align*}


\subsection{Selection Results and Analysis}
\label{sec.performance_llmselect}

As shown in~\Cref{tab:main_results}, \textit{AtS} outperforms baseline methods under both metrics and all budget ratios $\gamma$ on all datasets. While other methods might be good on one dataset but fail on another, \textit{AtS} performs consistently well. Even with access only to $\frac 1 {512}$ fraction of $\gS$, \textit{AtS} can capture the rank of the full-fine-tuning performance of different models with \textbf{PearCorr} equaling to $66.6\%$ in comparison to the second best method \textit{Zeroshot} with only $38.6\%$.  Our results also demonstrate the efficiency of \textit{AtS}. Indeed, it can select the model with averaged \textbf{RelAcc} larger than $95\%$ with $\gamma = \frac 1 {256}$, while all other methods fail to provide such a good selection even when $\gamma = \frac 1 8$. This implies that \textit{AtS} can select the near-optimal model with hundreds of times of acceleration.

\begin{figure}[t]
    \centering
    \includegraphics[width=0.9\linewidth]{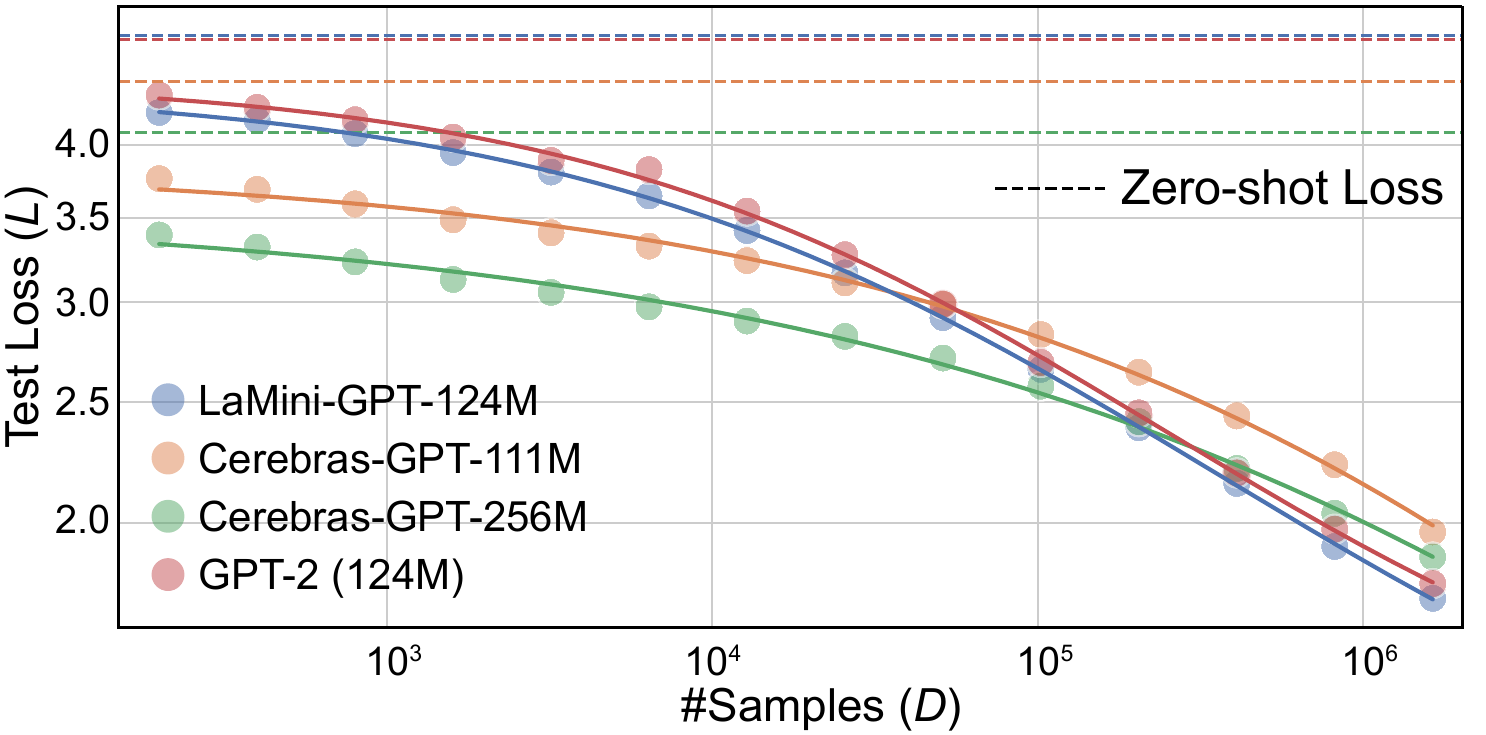}
    \vspace{-1em}
    \caption{Failure cases for the three baseline methods. The horizontal dashlines denote the zero-shot performance, and each point denotes the test loss when fine-tuning the corresponding model on $\gS_{sub}$ with size $D$. LaMini-GPT-124M has the best full-fine-tuning performance, but its performance on small $D$ is bad.}
    \label{fig:failure}
\end{figure}

\paragraph{Why do other algorithms fail?}
We illustrate why intuitively reasonable methods fail to make predictions in~\Cref{fig:failure}. Assume we have $4$ models and $|\gS_{sub}|$ is roughly $10^4$. \emph{ModelSize} selects the largest model in $\gM$ regardless of the properties of the downstream task and the models. The assumption behind this is that performance grows with model size, which has been demonstrated to be inaccurate in the fine-tuning stage. \emph{ZeroShot} and \emph{SubTuning} both leverage the performance on the downstream dataset. However, they only capture the performance under a specific dataset size, while ignoring the global trend of performance with data size. In fact, these methods give Cerebras-GPT-256M the highest score, but eventually, LaMini-GPT-124M outperforms.

\begin{figure*}[tp]
    \centering
    \includegraphics[width=0.95\linewidth]{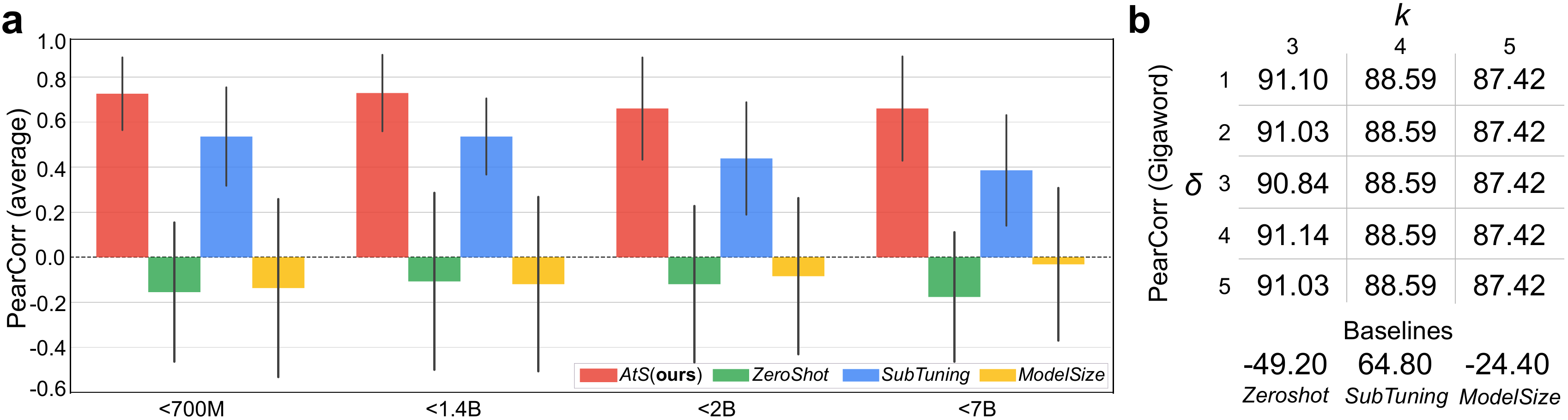}
    \caption{\textbf{(a)} \textbf{PearCorr} of \emph{AtS} on Gigaword with $\gamma=1/512$ under different memory budgets (different $\gM$). Full results are presented in Appendix \ref{app:ablation_size}. \textbf{(b)} Impact of $\delta$ and $k$ on \textbf{PearCorr}(\%) on Gigaword with $\gamma=1/512$. Full results are presented in Appendix \ref{app:ablation_hyper}.
    }
    \label{fig:ablation}
    \vspace{-5pt}
\end{figure*}

\paragraph{\textit{AtS} on stratified $\gM$.} 
We also consider different model sets $\gM$ to simulate the constraints of GPU memory. Specifically, we create three subsets of $\gM$ with different model size thresholds including $2B$, $1.4B$ and $700M$. The results are presented in \cref{fig:ablation}~(a), where \textit{AtS} outperforms other baselines on all subsets by a large margin.

\textbf{Influence of $k$ and $\delta$.}
To illustrate the influence of the outlier tolerance $\delta$ and the minimum accepted rate $k$, we conduct ablation studies on the choice of hyper-parameters and present the results in ~\Cref{fig:ablation}~(b). Overall, \emph{AtS} is not sensitive to hyper-parameters values, indicating its robustness under various circumstances.

\begin{table}[tb]
    \renewcommand{\arraystretch}{1}
    \centering
    \caption{\textbf{PearCorr}(\%) of three scaling-law-based selection methods on three datasets ($\gamma=1/512$). Full results are presented in Appendix \ref{app:ablation_different_laws}.}
\begin{tabular}{ccccc}
\toprule
 & FLAN & WMT19 & Gigaword & Avg. \\
\midrule


\textit{AtS} & \textbf{45.6} & \textbf{61.5} & \textbf{91.0} & \textbf{66.0} \\
\textit{OurFit} & \underline{36.8} & \textbf{61.5} & 78.5 & \underline{58.9} \\
\textit{VanillaFit} & 20.7 & 56.5 & \underline{79.3} & 52.1 \\

\bottomrule
\end{tabular}
    \label{tab:different_law_for_selection}
    \vspace{-8pt}
\end{table}

\textbf{LLM selection by fitting Scaling Law.} 
\textit{AtS} essentially leverages the proposed Scaling Law to estimate the trend of fine-tuning loss. Here we additionally consider two variants of using Scaling Laws: (1) \emph{OurFit} fine-tunes each model on a sequence of subsets $\{\gS_{sub}, \gS_{sub}^{1}, ...\}$ where $\gS_{sub}^{i}\subset \gS_{sub}, |\gS_{sub}^{i}|=2^{- i}|\gS_{sub}|$ until $|\gS_{sub}|<200$. It fits parameters in our law (\cref{eq.law}) using \textit{all} data-loss pairs, and predicts the performance on $\gS$ using the fitted law. (2) \emph{VanillaFit} follows a similar procedure, except that it fits the previous law (\cref{eq:prev_law_fix}) rather than ours. As shown in \cref{tab:different_law_for_selection}, while all variants outperform the three intuitive methods above, \textit{AtS} is better than \emph{OurFit} and \emph{VanillaFit} thanks to the robustness and stability brought by linearity.

\paragraph{Efficiency Analysis.}
We further analyze the efficiency of \textit{AtS} in comparison with other methods. According to \citet{kaplan2020scaling}, the computational cost $C$ measured in floating point operations (FLOPs) for training can be estimated with the formula $C \sim 6ND$, where $N$ represents the number of model parameters and $D$ the dataset size. Considering $T$ training epochs and $H$ hyper-parameter search rounds for each model on a given dataset, we estimate the overall computational costs for \textit{FullTuning}, \textit{SubTuning}, and \textit{AtS} as:
\begin{align*}
    C_\textit{FullTuning} &= \sum_{M\in\mathcal M}6N_M DTH\\
    C_\textit{SubTuning} &= \sum_{M\in\mathcal M}6N_M(\gamma D)TH = \gamma\cdot C_\textit{FullTuning} \\
    C_\textit{AtS} &= \sum_{M\in\mathcal M}\sum_{2^i\leq \gamma}6N_M\frac1{2^i} DTH\leq 2\gamma \cdot C_\textit{FullTuning} 
\end{align*}

\begin{figure*}[tb]
    \includegraphics[width=\linewidth]{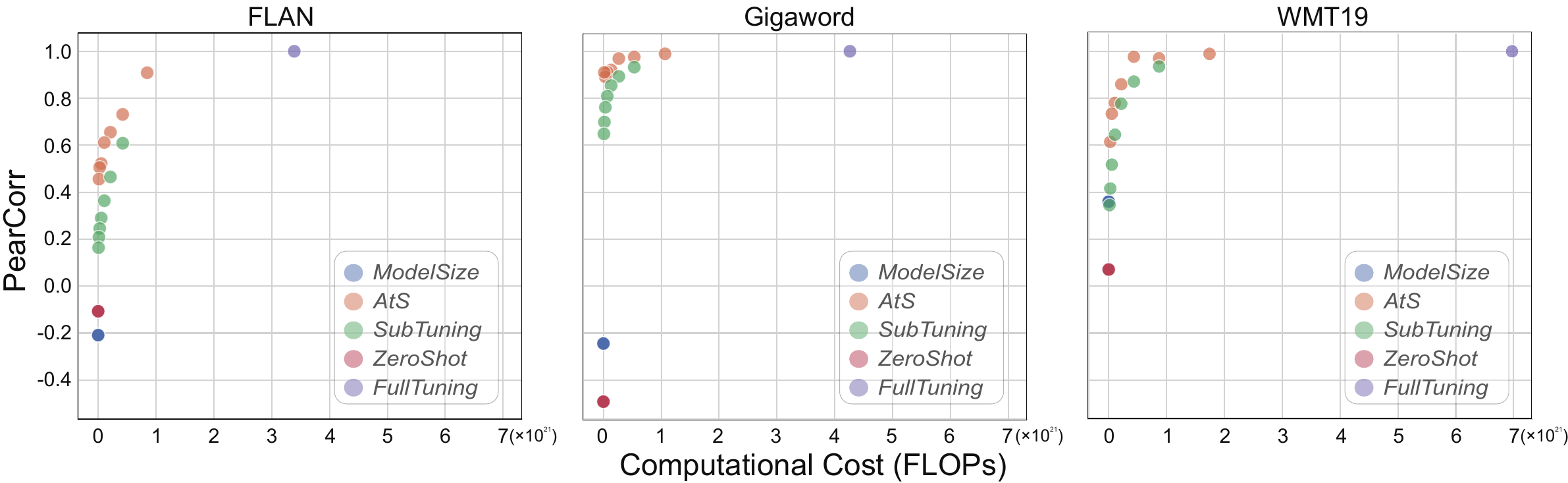}
    \vspace{-15pt}
    \caption{Pareto-Optimality curve between the selection performance and the computational costs. The performance is evaluated by \textbf{PearCorr} while the cost is evaluated by the number of floating point operations (FLOPs).}
    \label{fig:pareto}
\end{figure*}

Both \textit{AtS} and \textit{SubTuning} exhibit the same order of computational complexity, achieving an acceleration rate of $\gamma$. Figure \ref{fig:pareto} illustrates a Pareto-optimality curve between selection performance and computational costs. Notably, \textit{AtS} achieves the most optimal Pareto curve, providing near-optimal selection performance akin to \textit{FullTuning} while significantly reducing computational costs.

\section{Discussion}

\paragraph{Phase transition may happen on certain loss value.} As discussed in~\Cref{sec.ftlaw}, a more difficult downstream task results in a ``later'' occurrence of phase transition, which means more training samples are needed for the fine-tuned LLMs to enter the power phase. This phenomenon is justified by our results on FLAN, WMT19, and Gigaword (see~\cref{fig:sec3-main-law} and~\cref{app:fine-tune-results}). It is intuitive that the multi-task instruction tuning dataset FLAN is the most ``difficult'', followed by the machine translation dataset WMT19, and then the summarization task Gigaword. In addition, the stronger pre-trained LLMs will enter the power phase earlier. An interesting and unified explanation for this phenomenon is that, \emph{the phase transition may be closely related to the value of the test loss}. We also observed that \textbf{almost all models enter the power phase when their test loss is less than $2.2$}. This magic number is also reported in~\citet{du2024understanding}, where they found the emergent abilities of LLMs~\citep{schaeffer2024emergent} also emerge when the loss is smaller than $2.2$. The intrinsic mechanism behind this value still remains a mystery. It may suggest that there exist two distinct ``stages'' in general LLM learning process, which is similar but not identical to the Grokking of LLMs~\citep{liu2022towards}.

\paragraph{Limitations of this paper.}
Although \textit{AtS} can outperform other baselines significantly as shown in~\Cref{tab:main_results}, it also suffers performance degradation when data budget ratio $\gamma$ is extremely small, and all points we observed are in the pre-power phase. However, a mixed blessing is that in real applications, it is feasible to detect which stage the curve is in by monitoring the residual errors. Proposing a new algorithm that can make accurate predictions with observations only from the pre-power phase is an interesting direction to pursue. In addition, it will be interesting to see if the benefit of Scaling Laws can be extended to other fine-tuning strategies such as RLHF~\citep{Rafailov2023DirectPO,Christiano2017DeepRL}, LoRA~\citep{hu2021lora}, or more resource constraint types. Another limitation is a lack of a more comprehensive understanding of the mechanism of the pre-power phase and the phase transition. It will be interesting to see if it also appears under situations outside standard fine-tuning, and whether the behavior in this phase is similar to that in fine-tuning.

\paragraph{Outlook on Scaling Law research.}
We are now in a so-called ``post-LLM era'', where LLMs are revolutionizing various domains, such as human-like chatbot~\citep{team2023gemini}, clinical applications~\citep{Singhal2022LargeLM}, programming optimization~\citep{RomeraParedes2023MathematicalDF}, and geometric proofing~\citep{Trinh2024SolvingOG}. Scaling Law may be the key to unlocking the huge power of LLMs, since they tell us how can we make progress by investing more resources. However, research on Scaling Law is extremely expensive, and issues like environmental protection have to be concerned~\citep{Muennighoff2023ScalingDL}. We believe the research on this domain should be conducted in a \textit{collaborative and decentralized} manner, where the community can share the observed results and better utilize idle computational resources.


\section{Related Works}
\label{sec:relate+future}

\paragraph{Model selection.} 
Early model selection methods require that all models share identical architectures and differ only in pre-trained datasets~\citep{emd, nce}. Those similarity-based methods~\citep{Vu2020ExploringAP,Dwivedi2020DualityDS} fine-tune a model on a target dataset, and use the feature similarity between this model and candidate models to predict the fine-tuning performance for each model. \cite{ye2021towards} extends the feature-based method to model selection under the out-of-distribution setting. Another line of works design training-free metrics to examine whether pre-trained features are easily transferred to target tasks~\citep{Pandy2021TransferabilityEU,Ibrahim2021NewerIN}. More recently, there has been attempts to formulate the problem as learning to recommend~\citep{Li2023GuidedRF} or rank~\citep{Zhang2023ModelSL}. One reason for not adopting existing model selection methods outside LLM is that they focus mainly on classification or regression tasks~\citep{deshpande2021linearized,li2023guided}.
These methods either rely on features of inputs \citep{Lin2023ClassIL} or consider a fixed label set \citep{Nguyen2020LEEPAN}, which is not appropriate in the open-world text generation setting and could lead to the one-to-many problems \citep{Bao2019PLATOPD}.
The ever-growth of open-sourced LLM models urgently calls for the investigation of LLM selection. 

\paragraph{Scaling Law.} Laws between model performance and variables like model size or data size during pre-training have been widely studied~\citep{rosenfeld2019constructive,Aghajanyan2023ScalingLF,fernandes2023scaling,Frantar2023ScalingLF}, and are applied to estimate an optimal allocation of compute for pre-training LLMs~\citep{kaplan2020scaling,Hoffmann2022TrainingCL}. Recently, more fine-grained Scaling Laws have been proposed, such as data-constrained scaling~\citep{Muennighoff2023ScalingDL} and hyper-parameter scaling~\citep{Bi2024DeepSeekLS}. For LLM fine-tuning,~\citet{hernandez2021scaling} compared the scaling effect between transfer learning and pre-training, and~\citet{tay2021scale} observed the inconsistency of model size scaling between pre-training and fine-tuning. A concurrent work~\citep{zhang2024scaling} suggested a multiplicative law in fine-tuning scaling. However, none of these studies identified the pre-power phase in the fine-tuning process under low-data regimes, and their models fail to capture this phase transition pattern. 
Within the broader context of deep learning, \citet{rosenfeld2019constructive,alabdulmohsin2022revisiting,caballero2023broken} posited the necessity of a transition phase bridging the initial random-guess point and the power-law region in from-scratch training processes.  
Their primary approach involved modeling different phases separately and integrating them using a smooth function, which essentially introduced more parameters for Scaling Law.
In contrast, our proposed Rectified Scaling Law focuses on the fine-tuning of LLMs, and parameterizes the transition with a single term representing the pre-learned data size. This rectification is not only simple and intuitive but also empirically validated through solid experiments.

\section{Conclusion}
This paper focuses on two main areas: exploring the Scaling Laws of LLM fine-tuning and addressing the challenge of selecting LLMs for effective fine-tuning. We reveal the inadequacy of conventional Scaling Laws and propose a rectified law with much better theoretical and empirical properties by incorporating the concept of pre-learned data size. Additionally, we present a novel framework for the LLM selection problem and design a new algorithm that leverages the proposed law with significantly improved performance. Our findings not only deepen the understanding of Scaling Laws but also offer actionable insights for selecting LLMs in practice. We aim to provide a robust foundation for the broader and more efficient application of LLMs across various fields.

\section*{Acknowledgement}

This work is funded in part by the National Key R\&D Program of China \#2022ZD0160301, a grant from CCF-Tencent Rhino-Bird Open Research Fund. 

\section*{Impact Statement}

LLMs require huge amounts of computing power and energy to train and deploy, which results in carbon emissions and climate change. By designing a highly efficient algorithm to select LLMs for fine-tuning, our work significantly reduces the amount of time and resources required to achieve the best performance. This can lead to fewer energy consumption and lower cost, making LLMs more affordable and accessible to labs and start-ups when they have a certain task to solve. Our work is fundamental because it contributes to the development of more sustainable and responsible LLM selection process, which can have positive impact for the environment and society. Our method approaches a general problem and will not have any direct negative impact or be misused in specific domains as long as the task itself is safe, ethical and fair.

\bibliography{icml2024}
\bibliographystyle{icml2024}

\newpage
\appendix
\onecolumn

\newpage

\section{Fitting Scaling Laws: Optimization}
\label{app.optimization}

\subsection{Fitting of Vanilla Law}
Previous works \citep{kaplan2020scaling,hernandez2021scaling,tay2021scale} propose scaling laws sharing the following form:
\begin{align}
    \hat \gL(D) = \big(\frac{B}{D^\beta} + E\big)^{\alpha},
\end{align}
where $D$ is the number of training data, $B, E, \alpha, \beta$ are non-negative parameters that are model/task-dependent. Following \citet{Hoffmann2022TrainingCL}, we estimate $\{B, E, \alpha, \beta\}$ for each model by minimizing the following optimization problem,
\begin{align}
    \min_{B, E, \alpha, \beta}\sum_{\text{Run}\,i}\text{Huber}_\delta (\alpha\cdot \text{LSE}(\log B-\beta\log D_i, \log E) - \log \mathcal{L}_i)
\end{align}
where $\mathcal{L}_i$ denotes the test loss of fine-tuning on the data size $D_i$, \textit{LSE} denotes the log-exp-sum operator, \textit{Huber} denotes the Huber loss with $\delta=0.001$. We find the local minima of the objective above with the standard python package \textit{scipy} \citep{2020SciPy-NMeth} starting from $50$ random initialization of parameters. We choose the best one for reports.

\subsection{Fitting of Our Law}
Here we repeat the equation of our proposed fine-tuning scaling law for clarity:
\begin{align}
    \hat \gL(D) = \frac{B}{D_l+D^\beta} + E,
\end{align}
where $D_l$ represents the amount of data the model has pre-learned, $\beta$ denotes the learning difficulty, $B$ adjusts the initial test loss, and $E$ denotes the optimal loss of the model given an infinite amount of data. They are all model/task-dependent.
Similar with the fitting of vanilla law, we estimate $\{B, E, D_l, \beta\}$ for each model by minimizing the Huber loss,
\begin{align}
    \min_{B, E, \alpha, \beta}\sum_{\text{Run}\,i}\text{Huber}_\delta (\text{LSE}(\log B-\log(D_l+D^\beta), \log E) - \log \mathcal{L}_i)
\end{align}
We also repeat optimization for $50$ times and choose the best run for reports.

\subsection{Fit qualities of Vanilla Law and Our Law}

We fit both our law and the vanilla law on the fine-tuning performance of $30$ models (See Appendix \ref{app:fine-tune-results} for details). The root-mean-square deviation (RMSD) of fitted laws on each model is presented in Table \ref{app:rmsd_result}. The results 
demonstrates the superior fit quality of our proposed law over the vanilla law during the fine-tuning stage.

\begin{table*}
\centering
\caption{Comparison of root-mean-square deviation (RMSD) for fitting different scaling laws. $\Delta$ indicates the improvements on fitting quality of our proposed law over the vanilla law.}
\label{app:rmsd_result}
\begin{tabular}{l|ccc|ccc|ccc}
\hline
~ & \multicolumn{3}{c}{~~FLAN~~} & \multicolumn{3}{c}{~~WMT19~~} & \multicolumn{3}{c}{Gigaword} \\
\hline
Model Name & Ours & Vanilla & $\Delta$ & Ours & Vanilla & $\Delta$ & Ours & Vanilla & $\Delta$  \\
\hline
GPT2 & 0.0075 & 0.0697 & 0.0623 & 0.0089 & 0.1007 & 0.0918 & 0.0030 & 0.0190 & 0.0160 \\
GPT2-medium & 0.0038 & 0.0676 & 0.0639 & 0.0059 & 0.0991 & 0.0932 & 0.0020 & 0.0044 & 0.0024 \\
GPT2-large & 0.0056 & 0.0593 & 0.0537 & 0.0152 & 0.0893 & 0.0740 & 0.0035 & 0.0076 & 0.0041 \\
GPT2-xl & 0.0064 & 0.0614 & 0.0550 & 0.0410 & 0.1281 & 0.0871 & 0.0047 & 0.0104 & 0.0057 \\
LaMini-GPT-124M & 0.0027 & 0.0679 & 0.0652 & 0.0108 & 0.1150 & 0.1043 & 0.0037 & 0.0198 & 0.0161 \\
LaMini-GPT-774M & 0.0054 & 0.0638 & 0.0584 & 0.0093 & 0.1074 & 0.0981 & 0.0019 & 0.0074 & 0.0055 \\
LaMini-GPT-1.5B & 0.0055 & 0.0664 & 0.0609 & 0.0150 & 0.1353 & 0.1202 & 0.0063 & 0.0104 & 0.0041 \\
Cerebras-GPT-111M & 0.0096 & 0.0601 & 0.0505 & 0.0098 & 0.1129 & 0.1032 & 0.0038 & 0.0219 & 0.0181 \\
Cerebras-GPT-256M & 0.0105 & 0.0517 & 0.0412 & 0.0095 & 0.0874 & 0.0780 & 0.0022 & 0.0137 & 0.0115 \\
Cerebras-GPT-1.3B & 0.0038 & 0.0188 & 0.0150 & 0.0131 & 0.0618 & 0.0488 & 0.0048 & 0.0159 & 0.0111 \\
Cerebras-GPT-2.7B & 0.0030 & 0.0033 & 0.0003 & 0.0114 & 0.0123 & 0.0009 & 0.0020 & 0.0022 & 0.0002 \\
Phi-1.5 & 0.0112 & 0.0363 & 0.0251 & 0.0110 & 0.0366 & 0.0256 & 0.0029 & 0.0030 & 0.0001 \\
Phi-2 & 0.0060 & 0.0197 & 0.0137 & 0.0101 & 0.0317 & 0.0216 & 0.0040 & 0.0049 & 0.0008 \\
OPT-350m & 0.0078 & 0.0478 & 0.0400 & 0.0135 & 0.0848 & 0.0712 & 0.0045 & 0.0055 & 0.0010 \\
OPT-1.3b & 0.0024 & 0.0165 & 0.0141 & 0.0150 & 0.0709 & 0.0558 & 0.0024 & 0.0034 & 0.0010 \\
OPT-2.7b & 0.0052 & 0.0072 & 0.0020 & 0.0229 & 0.0602 & 0.0373 & 0.0012 & 0.0018 & 0.0006 \\
OPT-6.7b & 0.0025 & 0.0026 & 0.0002 & 0.0073 & 0.0090 & 0.0016 & 0.0028 & 0.0030 & 0.0002 \\
ai-forever/mGPT & 0.0035 & 0.0050 & 0.0015 & 0.0049 & 0.0153 & 0.0104 & 0.0119 & 0.0119 & 0.0000 \\
BART-base & 0.0073 & 0.0506 & 0.0433 & 0.0201 & 0.1075 & 0.0873 & 0.0194 & 0.0247 & 0.0053 \\
BART-large & 0.0129 & 0.0388 & 0.0259 & 0.0123 & 0.1070 & 0.0947 & 0.0055 & 0.0054 & -0.0001 \\
BART-large-cnn & 0.0115 & 0.0302 & 0.0187 & 0.0115 & 0.0747 & 0.0632 & 0.0053 & 0.0059 & 0.0006 \\
BART-large-xsum & 0.0090 & 0.0357 & 0.0267 & 0.0089 & 0.1011 & 0.0922 & 0.0039 & 0.0046 & 0.0006 \\
T5-small & 0.0039 & 0.0241 & 0.0202 & 0.0135 & 0.0141 & 0.0007 & 0.0079 & 0.0235 & 0.0156 \\
T5-base & 0.0078 & 0.0316 & 0.0238 & 0.0144 & 0.0151 & 0.0007 & 0.0026 & 0.0134 & 0.0108 \\
mT5-base & 0.0035 & 0.0136 & 0.0101 & 0.0066 & 0.0155 & 0.0088 & 0.0055 & 0.0277 & 0.0221 \\
mT5-large & 0.0027 & 0.0118 & 0.0091 & 0.0045 & 0.0249 & 0.0204 & 0.0024 & 0.0071 & 0.0046 \\
T5-v1.1-base & 0.0069 & 0.0456 & 0.0386 & 0.0117 & 0.0358 & 0.0241 & 0.0056 & 0.0056 & 0.0000 \\
switch-base-8 & 0.0073 & 0.0298 & 0.0225 & 0.0098 & 0.0104 & 0.0006 & 0.0096 & 0.0110 & 0.0014 \\
switch-base-16 & 0.0088 & 0.0284 & 0.0195 & 0.0154 & 0.0171 & 0.0017 & 0.0082 & 0.0074 & -0.0008 \\
switch-base-32 & 0.0103 & 0.0307 & 0.0204 & 0.0048 & 0.0058 & 0.0009 & 0.0109 & 0.0131 & 0.0022 \\
\hline
\end{tabular}
\end{table*}

\newpage
\section{Details of Studied LLMs}
\label{app.llm}

\begin{table*}[th]
\caption{This table summarizes all the models we used in experiments. The Arch. is short for model architecture, De-only, En-De and Moe stands for Decoder-only, Encoder-Decoder and Mixture of Experts respectively. The last few columns summarize the configuration of different language models, including number of parameters, number of layers, dimension of hidden states, number of attention heads, dimension of feed-forward layers, and dimension of key/value head.}
\resizebox{\linewidth}{!}{
\begin{tabular}{c|c|c|c|c|c|c|c|c}
\hline
Model Name & Arch. & Training Data Source & $N$ & $N_{layer}$ & $d_{model}$ & $N_{head}$ & $d_{ff}$ & $d_{kv}$ \\ \hline
GPT-2 & \multirow{21}{*}{De-only} & \multirow{4}{*}{WebText} & 124M & 12 & 768 & 12 & 3072 & 64 \\
GPT-2-medium &  &  & 354M & 24 & 1024 & 16 & 4096 & 64 \\
GPT-2-large &  &  & 774M & 36 & 1280 & 20 & 5120 & 64 \\
GPT-2-xl &  &  & 1.5B & 48 & 1600 & 25 & 6400 & 64 \\ \cline{1-1} \cline{3-9} 
LaMini-GPT-124M &  & \multirow{3}{*}{Finetuned GPT-2-XL} & 124M & 12 & 768 & 12 & 3072 & 64 \\
LaMini-GPT-774M &  &  & 774M & 36 & 1280 & 20 & 5120 & 64 \\
LaMini-GPT-1.5B &  &  & 1.5B & 48 & 1600 & 25 & 6400 & 64 \\ \cline{1-1} \cline{3-9} 
Cerebras-GPT-111M &  & \multirow{4}{*}{The Pile} & 111M & 10 & 768 & 12 & 3072 & 64 \\
Cerebras-GPT-256M &  &  & 256M & 14 & 1088 & 17 & 4352 & 64 \\
Cerebras-GPT-1.3B &  &  & 1.3B & 24 & 2048 & 16 & 8192 & 128 \\
Cerebras-GPT-2.7B &  &  & 2.7B & 32 & 2560 & 32 & 10240 & 80 \\ \cline{1-1} \cline{3-9} 
Phi-1.5 &  & \multirow{2}{*}{Mixed Real \& Synthetic Data} & 1.4B & 24 & 2048 & 32 & 8192 & 64 \\
Phi-2 &  &  & 2.7B & 32 & 2560 & 32 & 10240 & 80 \\ \cline{1-1} \cline{3-9} 
OPT-350m &  & \multirow{4}{*}{\begin{tabular}[c]{@{}c@{}}BookCorpus, CC-Stories, \\ The Pile, Pushshift.io, CCNewsV2\end{tabular}} & 331M & 24 & 1024 & 16 & 4096 & 64 \\
OPT-1.3b &  &  & 1.3B & 24 & 2048 & 32 & 8192 & 64 \\
OPT-2.7b &  &  & 2.7B & 32 & 2560 & 32 & 10240 & 80 \\
OPT-6.7b &  &  & 6.7B & 32 & 4096 & 32 & 16384 & 128 \\ \cline{1-1} \cline{3-9} 
ai-forever/mGPT &  & Multilingual Wikipedia and C4 & 1.4B & 24 & 2048 & 16 & 8192 & 128 \\ \hline
BART-base & \multirow{9}{*}{En-De} & \multirow{2}{*}{\begin{tabular}[c]{@{}c@{}}BookCorpus, CCNews, \\ OpenWebText, STORIES\end{tabular}} & 96M & 12/12 & 768 & 12 & 3072 & 64 \\
BART-large &  &  & 254M & 12/12 & 1024 & 16 & 4096 & 64 \\ \cline{1-1} \cline{3-9} 
BART-large-CNN &  & BART finetuned on CNN & 254M & 12/13 & 1024 & 16 & 4096 & 64 \\ \cline{1-1} \cline{3-9} 
BART-large-XSUM &  & BART finetuned on XSUM & 254M & 12/14 & 1024 & 16 & 4096 & 64 \\ \cline{1-1} \cline{3-9} 
T5-small &  & \multirow{2}{*}{\begin{tabular}[c]{@{}c@{}}C4, Wiki-DPR, finetuned on CoLA, SST-2, \\ MRPC, STS-B, QQP, MNLI, QNLI etc.\end{tabular}} & 60M & 6/6 & 512 & 8 & 2048 & 64 \\
T5-base &  &  & 223M & 12/12 & 768 & 12 & 3072 & 64 \\ \cline{1-1} \cline{3-9} 
mT5-base &  & \multirow{2}{*}{mC4} & 582M & 12/12 & 768 & 12 & 2048 & 64 \\
mT5-large &  &  & 1.2B & 24/24 & 768 & 12 & 2816 & 64 \\ \cline{1-1} \cline{3-9} 
T5-v1.1-base &  & C4 & 247M & 12/12 & 768 & 12 & 2048 & 64 \\ \hline
switch-base-32 & \multirow{3}{*}{En-De MoE} & \multirow{3}{*}{C4} & 2B & 12/12 & 768 & 12 & 3072 & 64 \\
switch-base-16 &  &  & 1B & 12/12 & 768 & 12 & 3072 & 64 \\
switch-base-8 &  &  & 619M & 12/12 & 768 & 12 & 3072 & 64
\end{tabular}}

\end{table*}

\paragraph{GPT-2 Series~\citep{Radford2019LanguageMA}}
GPT-2 series are transformer-based language models created and released by OpenAI. The models are pre-trained on WebText with 40GB of English text that is not publicly released. The texts are tokenized using a byte-level version of Byte Pair Encoding (BPE) and a vocabulary size of 50,257. The pre-training objective is causal language modeling (CLM). In this paper, we studied all the released versions of GPT-2, which includes GPT2 (124M), GPT2-Medium (355M), GPT2-Large (774M), and GPT2-XL (1.5B).

\paragraph{OPT Series~\citep{Zhang2022OPTOP}}
Open Pre-trained Transformers (OPT) is a suite of decoder-only pre-trained transformers released on May 3rd 2022 by Meta AI. OPT was predominantly pre-trained with English text, but a small amount of non-English data is present within the training corpus via CommonCrawl. The training data of OPT contains 180 tokens corresponding to 800GB of data, which is composed of texts from BookCorpus, CC-Stories, The Pile, Pushshift.io Reddit, and CCNewsV2. The texts are tokenized using the GPT2 byte-level version of BPE and a vocabulary size of 50,272. In this paper, we studied 5 versions of OPT, including OPT-350M, OPT-1.3B, OPT-2.7B, and OPT-6.7B.

\paragraph{Phi Series~\citep{Li2023TextbooksAA}}
Phi are transformer-based language models created and released by Microsoft to investigate the ability of smaller models. Their main goal is to answer ``how small can a LLM be to achieve certain capabilities". Its training involved a variety of data sources related to code produced by humans and LLMs. Phi series includes 3 pre-trained models without fine-tuning or RLHF: Phi-1 (1.3B), Phi-1.5 (1.3B), and Phi-2 (2.7B). They have shown nearly state-of-the-art performance among models much larger than them. In this paper, we studied Phi-1.5 and Phi-2.

\paragraph{LaMini-LM Series~\citep{wu2023lamini}}
To alleviate the resource-intensive problem, \citet{wu2023lamini} explored new ways of distilling knowledge from large models into smaller ones. They designed a new pipeline that combines synthetic data with existing instructions to produce a wide variety of instruction training datasets consisting of over 2.58 million examples. Based on these instructions, they finetuned a diverse herd of language models including encoder-decoder and decoder-only families and named them ``LaMini-LMs", with parameters ranging from 61M to 1.5B. We chose the LaMiniGPT series in our experiments, which are some of the largest models available in the LaMini family.

\paragraph{Cerebras-GPT~\citep{dey2023cerebras}}

The cerebras-GPT family is inspired by the Chinchilla Scaling laws which state that a ratio of 20 training tokens per model parameter is optimal for computational cost. These models share similar architecture to GPT-3, but only pre-trained on The Pile. Cerebras-GPTs use Byte Pair Encoding and have a vocabulary of 50257 words. In this paper, we studied Cerebras-GPT-111M, Cerebras-GPT-256M, Cerebras-GPT-1.3B, and Cerebras-GPT-2.7B.

\paragraph{T5, T5\_V1.1 and mT5 Series~\citep{raffel2020exploring,xue2020mt5}}

T5(text-to-text transfer Transformers) is an encoder-decoder language model, first introduced in~\citet{raffel2020exploring}. T5 was pre-trained on C4 and fine-tuned on several downstream datasets, which achieved state-of-the-art on many benchmarks including question answering, text classification, and machine translation.
T5-V1.1 shares a similar architecture with T5, except for adopting GeGLU as nonlinearities and scaling down both $d_{model}$ and $d_{ff}$. In contrast to T5, T5-V1.1 was only pre-trained on C4.
mT5 is a multilingual variant of t5-V1.1 which was pre-trained on unlabeled multilingual Common-Crawl~(mC4) dataset without dropout. mT5's training corpus consisted of 101 languages, which makes it directly applicable to multilingual settings. We chose T5-small, T5-base, T5-V1.1-base, mT5-base and mT5-large in our experiments.

\paragraph {BART Series~\citep{lewis2019bart}}

BART is a sequence-to-sequence model with a bidirectional encoder and an auto-regressive decoder. It was trained by two steps: (1) introducing noise to the pre-train text with an arbitrary function, and (2) learning to reconstruct the original text. BART was trained on a mixture of corpora consisting of BookCorpus, CCNews, OpenWebText, and STORIES. In this work, we chose BART-base, BART-large, BART-large-CNN, and BART-large-xsum for experiments. The last two models are BART-large finetuned on CNN and XSUM datasets respectively, making them suitable for text summary tasks.

\newpage

\section{Details of Datasets}
\label{app.datasets}
We mainly conducted experiments on three datasets, WMT-19, Gigaword, and FLAN. The first two tasks~(Machine Translation and Summarization) are traditional sequence-to-sequence NLP tasks. The FLAN dataset consists of different generation tasks in many formats, which is an ideal benchmark for evaluating LLMs' performance in day-to-day situations. The statistics of the three datasets are shown in \cref{tab:dataset_statistics}\footnote{We re-partition datasets into train/validation/test subsets due to the unavailability of the WMT19 test set and the imbalance in the split between the validation and test sets within Gigaword. We only sub-sample a subset from FLAN since the full dataset is too large.}.

\paragraph{FLAN~\citep{longpre2023flan}}
The Flan Collection consolidates datasets from Flan 2021, P3, Super-Natural Instructions, and dozens of others into a single repository. It then formats them into a variety of templates, including zero-shot, few-shot, and chain-of-thought formats. In our experiments, we use the FLAN Collection provided by Huggingface~\footnote{https://huggingface.co/datasets/Open-Orca/FLAN} and we choose the no-option split which requires the model to generate a free-form answer.

\paragraph{WMT19~\citep{wmt19translate}}
WMT-19 is a public machine translation dataset commonly used for evaluating sequence-to-sequence models. We initiated our experiments on WMT-19 En-Zh.
Considering the instruction-tuned models within our model set (e.g. LaMini-GPTs), we prepend an additional instruction ``Translate to Chinese:" at the beginning during fine-tuning.

\paragraph{Gigaword~\citep{graff2003english}}
Gigaword is a widely used resource in the field of text summarization, comprising billions of words from a vast collection of news articles like the New York Times and the Associated Press. Each news document in the dataset is paired with a professionally written headline, serving as a compact summary of the main ideas within the article. We also prepend an additional instruction ``Generate a summary: " to input sequences in the dataset.

\begin{table}[htb]
    \centering
    \caption{Statistics of fine-tuning datasets}
    \begin{tabular}{c|c|c|c}
        \hline
        Dataset & Input length (Avg/Max) & Target length (Avg/Max) & Dataset Size (Train/Valid/Test) \\
        \hline
        FLAN & 23~/~117 & 12~/~96 & 2,320,656~/~10,000~/~10,000 \\
        WMT19  & 32~/~249 & 40~/~446  & 25,982,455~/~3,981~/~3,981  \\
        Gigaword & 36~/~70 & 8~/~19 & 3,795,957~/~8,000~/~8,000 \\
        \hline
    \end{tabular}
    \label{tab:dataset_statistics}
\end{table}

\makeatletter
\newcommand{\DrawLine}{%
  \begin{tikzpicture}
  \path[use as bounding box] (0,0) -- (\linewidth,0);
  \draw[color=black!75!black,dashed,dash phase=2pt]
        (0-\kvtcb@leftlower-\kvtcb@boxsep,0)--
        (\linewidth+\kvtcb@rightlower+\kvtcb@boxsep,0);
  \end{tikzpicture}%
  }
\makeatother

\begin{tcolorbox}[title = {Examples from FLAN}]

\textbf{Input}: What is the solution? Solve 134*c - 143 + 2957 = 0 for c. \\
\textbf{Target}: -21\\
\DrawLine \\
\textbf{Input}: Translate the following sentence to Czech: Let us finish it.\\
\textbf{Target}: Dokončeme to.\\
\DrawLine \\
\textbf{Input}: \\
Premise: Our world has what is for them a normal gravity, but because of our much higher gravitational potential, our atmosphere is too dense to support them comfortably over sustained periods. \\
Hypothesis: Your world has the same type of gravity as theirs. \\
Does the premise entail the hypothesis? \\
\textbf{Target}: Yes. \\
\DrawLine \\
\textbf{Input}: \\
How are binary trees extended? \\
How do I insert a new node on a binary tree (not search binary tree)? \\
Do those questions have the same meaning? \\
\textbf{Target}: no 
\end{tcolorbox}

\begin{CJK*}{UTF8}{gkai}

\begin{tcolorbox}[title = {Examples from WMT19}]
\textbf{Input}: Translate to Chinese: When the mother sheep saw him pick up her baby sheep and ran away, she followed him out of the field. \\
\textbf{Target}: {当羊妈妈看见她的羊宝宝被人抱走了，赶快跟在李雷后面跑出了田地。}\\
\DrawLine \\
\textbf{Input}: Translate to Chinese: South Africa's Draft White Paper on Energy Policy promotes energy efficiency and use of renewable sources of energy.\\
\textbf{Target}: 南非的《能源政策白皮书草案》提倡提高能源效率和使用可再生能源。\\
\DrawLine \\
\textbf{Input}: Translate to Chinese: Political scientists like Janine Mossuz-Lavau says there is being a woman this election season may be an asset.\\
\textbf{Target}: 政治学家如詹南·摩萨斯－拉瓦说，在这季奄中，身为女性也许就是资本。 \\
\DrawLine \\
\textbf{Input}: Translate to Chinese: The Secretary-General condemned the excessive and disproportionate use of force and the killing of civilians.\\
Target: 秘书长谴责这种不成比例地过度使用武力和杀害平民的行为。\\

\end{tcolorbox}
\end{CJK*}

\begin{tcolorbox}[title = {Examples from Gigaword}]
\textbf{Input}: Generate a summary: china is to hold the third international expo of necessities for students in nanning city in south china 's guangxi zhuang autonomous region from october to november. \\
\textbf{Target}: china to hold expo of student equipment \\
\DrawLine \\
\textbf{Input}: Generate a summary: the gold price in hong kong rose \#\# hk dollars on wednesday to close at \#,\#\#\# hk dollars a tael , according to po sang bank , one of the major gold dealers in hong kong. \\
\textbf{Target}: gold price in hong kong up \\
\DrawLine \\
\textbf{Input}: Generate a summary: riot police used water cannons friday to disperse protesters demanding that the philippines lift its ban on the deployment of workers to war-ravaged iraq .\\
\textbf{Target}: police violently disperse protest against ban on workers deployment to iraq \\
\DrawLine \\
\textbf{Input}: Generate a summary: british prime minister john major thursday hailed the re-election of russian president boris yeltsin as a sign that `` democracy has taken firm root in russia .\\
\textbf{Target}: major delighted over yeltsin victory

\end{tcolorbox}

\newpage

\section{Details of Fine-tuning Experiments}
\label{app.ftsetting}

\subsection{Implementation Details}
We continue training each model initialized from the pretrained checkpoint with the standard cross-entropy loss on each target token. 
For decoder-only models, we concatenate the input sequence and the target sequence together through the decoder. For encoder-decoder models, we forward the input sequence and the target sequence through the encoder and the decoder respectively. The cross-entropy loss is calculated over the target tokens.

To ensure the best fine-tuning performance without interference from the choice of hyper-parameters, we conduct hyper-parameter searching for important ones including learning rate and batch size. We also conduct each experiment with the searched hyper-parameters three times and report the average performance.
All the experiments are implemented using transformers package \citep{wolf-etal-2020-transformers}. 

\begin{table}[htb]
    \centering
    \begin{tabular}{c|c}
        \hline
        Hyper-parameter & Values \\
        \hline
        learning rate &  \multicolumn{1}{c}{\begin{tabular}{@{}r@{}c@{}} search on &$~\{1e-4, 3e-4, 5e-4, 1e-3\}$ for small models $<700M$, \\  &$\{3e-5, 5e-5, 1e-4, 3e-4\}$ for large models $>700M$\end{tabular}} \\
        \hline
        batch size &  search on $\{64, 128, 256\}$ \\
        \hline
        training epoch &  $20$ with early stopping (patience=3) \\
        \hline
        optimizer & AdamW \\
        \hline
        weight decay & $0.01$ \\
        \hline
        scheduler & cosine \\
        \hline 
        warmup ratio & $0.03$\\
        \hline
        
    \end{tabular}
    \caption{Hyper-parameter settings of fine-tuning experiments.}
\end{table}

\subsection{Hardware and Software}
We run most of the experiments on clusters using NVIDIA A100s. We implemented our experiments using PyTorch~\citep{paszke2017automatic} and the HuggingFace library. For each model, we randomly sampled seeds for 3 runs and controlled the number of training samples. The total vocabulary size and tokenizer used varied from case to case. Overall, we estimated that a total of 20,000 GPU hours were consumed.

\newpage

\section{Results of Fine-tuning Experiments}
\label{app:fine-tune-results}

Here we present the experimental results of $30$ models fine-tuned on various sizes of subsets from WMT19, Gigaword, and FLAN. The subsets are randomly sampled from the original datasets. We repeat each experiment for three times with different random seeds and report the average. The fine tuning processes are very stable, and the variance is low. We report the variance of fine tuning results of four typical models on FLAN in Table \ref{tab:variance}.

\begin{table*}[htb]
    \centering
    \small
    \captionsetup{justification=centering}
    \caption{Test loss of $30$ models fine-tuned on subsets of FLAN dataset. The data size ranges from $0$ to $1638400$.}
\resizebox{\linewidth}{!}{
\begin{tabular}{lccccccccccccccc}
\toprule
\multicolumn{1}{c}{Model} & ~~~~0~~~~ & ~~~200~~~ & ~~~400~~~ & ~~~800~~~ & ~~~1600~~~ & ~~~3200~~~ & ~~~6400~~~ & ~~12800~~ & ~~25600~~ & ~~51200~~ & 102400 & 204800 & 409600 & 819200 & 1638400 \\
\midrule
GPT-2 & 4.857 & 4.386 & 4.288 & 4.191 & 4.060 & 3.890 & 3.826 & 3.546 & 3.272 & 2.988 & 2.686 & 2.449 & 2.193 & 1.978 & 1.791 \\
GPT-2-medium & 4.375 & 3.782 & 3.714 & 3.614 & 3.518 & 3.390 & 3.249 & 3.076 & 2.880 & 2.673 & 2.428 & 2.207 & 1.966 & 1.771 & 1.610 \\
GPT-2-large & 4.165 & 3.525 & 3.493 & 3.412 & 3.285 & 3.157 & 3.044 & 2.898 & 2.736 & 2.543 & 2.324 & 2.115 & 1.913 & 1.739 & 1.601 \\
GPT-2-xl & 3.929 & 3.306 & 3.254 & 3.169 & 3.058 & 2.999 & 2.889 & 2.774 & 2.632 & 2.451 & 2.270 & 2.058 & 1.878 & 1.693 & 1.555 \\
LaMini-GPT-124M & 4.891 & 4.248 & 4.188 & 4.087 & 3.946 & 3.808 & 3.645 & 3.421 & 3.165 & 2.916 & 2.653 & 2.383 & 2.152 & 1.917 & 1.743 \\
LaMini-GPT-774M & 4.215 & 3.497 & 3.458 & 3.361 & 3.257 & 3.140 & 3.033 & 2.878 & 2.712 & 2.529 & 2.329 & 2.120 & 1.887 & 1.731 & 1.559 \\
LaMini-GPT-1.5B & 4.046 & 3.293 & 3.240 & 3.202 & 3.094 & 2.990 & 2.881 & 2.751 & 2.628 & 2.446 & 2.270 & 2.061 & 1.851 & 1.687 & 1.530 \\
Cerebras-GPT-111M & 4.495 & 3.763 & 3.689 & 3.593 & 3.489 & 3.407 & 3.325 & 3.237 & 3.108 & 2.991 & 2.827 & 2.638 & 2.435 & 2.226 & 1.968 \\
Cerebras-GPT-256M & 4.097 & 3.393 & 3.319 & 3.230 & 3.127 & 3.054 & 2.974 & 2.898 & 2.817 & 2.708 & 2.572 & 2.409 & 2.211 & 2.037 & 1.880 \\
Cerebras-GPT-1.3B & 3.388 & 2.791 & 2.713 & 2.646 & 2.587 & 2.492 & 2.412 & 2.325 & 2.243 & 2.131 & 2.042 & 1.960 & 1.881 & 1.786 & 1.683 \\
Cerebras-GPT-2.7B & 2.914 & 2.231 & 2.151 & 2.088 & 2.046 & 1.979 & 1.925 & 1.872 & 1.831 & 1.779 & 1.733 & 1.681 & 1.631 & 1.589 & 1.544 \\
Phi-1.5 & 4.620 & 4.063 & 3.929 & 3.664 & 3.462 & 3.213 & 3.056 & 2.895 & 2.686 & 2.463 & 2.237 & 2.022 & 1.831 & 1.671 & 1.542 \\
Phi-2 & 3.368 & 2.538 & 2.515 & 2.452 & 2.424 & 2.397 & 2.386 & 2.330 & 2.292 & 2.216 & 2.146 & 2.076 & 2.009 & 1.944 & 1.882 \\
OPT-350m & 3.729 & 3.203 & 3.132 & 3.020 & 2.943 & 2.848 & 2.767 & 2.686 & 2.577 & 2.453 & 2.292 & 2.131 & 1.964 & 1.805 & 1.663 \\
OPT-1.3b & 3.022 & 2.447 & 2.379 & 2.317 & 2.268 & 2.189 & 2.110 & 2.042 & 1.973 & 1.902 & 1.821 & 1.742 & 1.672 & 1.596 & 1.513 \\
OPT-2.7b & 2.793 & 2.337 & 2.287 & 2.240 & 2.170 & 2.109 & 2.031 & 1.953 & 1.917 & 1.873 & 1.800 & 1.746 & 1.689 & 1.635 & 1.579 \\
OPT-6.7b & 4.442 & 2.021 & 1.980 & 1.973 & 1.935 & 1.921 & 1.895 & 1.865 & 1.838 & 1.812 & 1.790 & 1.770 & 1.741 & 1.720 & 1.697 \\
ai-forever/mGPT & 3.227 & 2.623 & 2.587 & 2.512 & 2.478 & 2.391 & 2.339 & 2.292 & 2.215 & 2.150 & 2.096 & 2.051 & 1.989 & 1.942 & 1.894 \\
BART-base & 8.502 & 4.159 & 3.990 & 3.850 & 3.685 & 3.532 & 3.344 & 3.181 & 2.979 & 2.711 & 2.457 & 2.251 & 2.051 & 1.858 & 1.685 \\
BART-large & 7.533 & 3.372 & 3.328 & 3.106 & 2.950 & 2.827 & 2.712 & 2.617 & 2.500 & 2.337 & 2.172 & 2.006 & 1.853 & 1.688 & 1.550 \\
BART-large-cnn & 6.026 & 3.591 & 3.445 & 3.213 & 3.037 & 2.894 & 2.757 & 2.606 & 2.471 & 2.338 & 2.164 & 1.999 & 1.829 & 1.674 & 1.555 \\
BART-large-xsum & 4.908 & 3.493 & 3.335 & 3.168 & 3.023 & 2.893 & 2.755 & 2.627 & 2.476 & 2.350 & 2.171 & 2.008 & 1.836 & 1.677 & 1.557 \\
T5-small & 3.983 & 3.021 & 2.931 & 2.838 & 2.757 & 2.681 & 2.601 & 2.508 & 2.411 & 2.309 & 2.208 & 2.085 & 1.978 & 1.857 & 1.756 \\
T5-base & 3.539 & 2.642 & 2.585 & 2.480 & 2.412 & 2.344 & 2.281 & 2.201 & 2.131 & 2.041 & 1.947 & 1.837 & 1.715 & 1.600 & 1.520 \\
mT5-base & 12.925 & 3.191 & 3.121 & 3.010 & 2.892 & 2.758 & 2.656 & 2.514 & 2.413 & 2.308 & 2.178 & 2.069 & 1.969 & 1.879 & 1.799 \\
mT5-large & 20.843 & 2.596 & 2.528 & 2.470 & 2.389 & 2.311 & 2.220 & 2.138 & 2.051 & 1.966 & 1.890 & 1.810 & 1.741 & 1.675 & 1.601 \\
T5-v1.1-base & 28.836 & 4.012 & 3.891 & 3.723 & 3.503 & 3.312 & 3.101 & 2.903 & 2.727 & 2.525 & 2.328 & 2.119 & 1.930 & 1.727 & 1.528 \\
switch-base-8 & 29.484 & 4.129 & 3.892 & 3.689 & 3.469 & 3.285 & 3.132 & 2.896 & 2.728 & 2.536 & 2.368 & 2.168 & 1.988 & 1.799 & 1.654 \\
switch-base-16 & 18.770 & 3.812 & 3.620 & 3.451 & 3.290 & 3.101 & 2.919 & 2.796 & 2.633 & 2.497 & 2.329 & 2.163 & 2.000 & 1.817 & 1.684 \\
switch-base-32 & 24.522 & 3.652 & 3.502 & 3.312 & 3.181 & 3.014 & 2.836 & 2.704 & 2.572 & 2.434 & 2.304 & 2.116 & 1.950 & 1.780 & 1.650 \\
\bottomrule
\end{tabular}
}
\end{table*}

\begin{figure}[!hbp]
    \centering
    \includegraphics[width=0.7\linewidth]{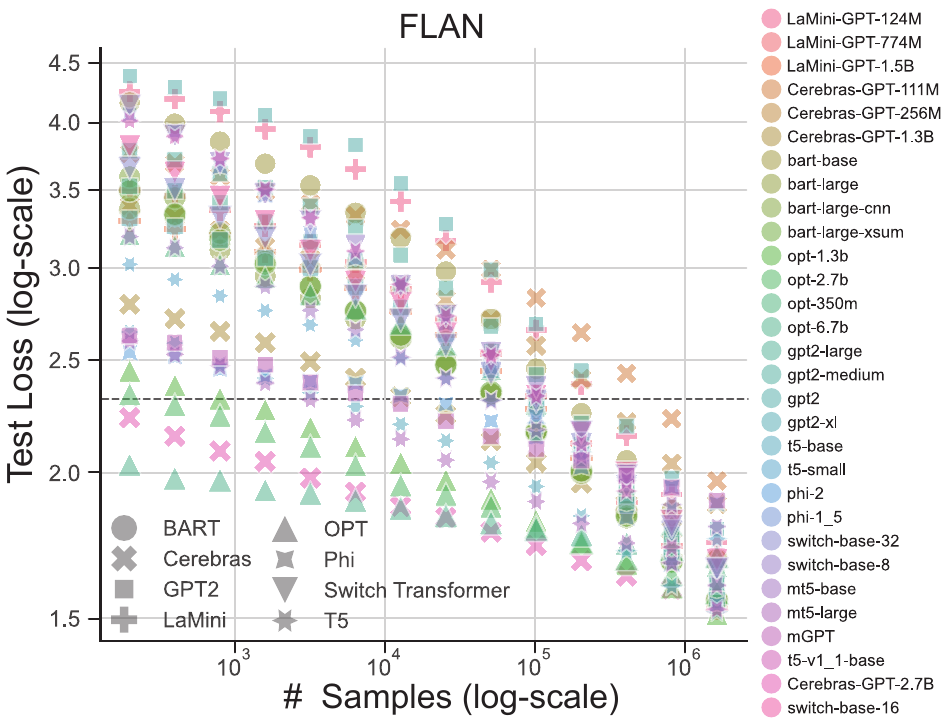}
    \label{fig:flan}
    \caption{The test losses of $30$ models fine-tuned on various sizes of subsets derived from FLAN dataset. The point size reflects the corresponding model size.}
\end{figure}

\newpage
\begin{table*}[htb]
    \centering
    \small
    \captionsetup{justification=centering}
    \caption{Test loss of $30$ models fine-tuned on subsets of WMT19 dataset. The data size ranges from $0$ to $1638400$.}
    \resizebox{\linewidth}{!}{
\begin{tabular}{lccccccccccccccc}
\toprule
\multicolumn{1}{c}{Model} & ~~~~0~~~~ & ~~~200~~~ & ~~~400~~~ & ~~~800~~~ & ~~~1600~~~ & ~~~3200~~~ & ~~~6400~~~ & ~~12800~~ & ~~25600~~ & ~~51200~~ & 102400 & 204800 & 409600 & 819200 & 1638400 \\
\midrule
GPT-2 & 3.403 & 3.079 & 3.037 & 2.955 & 2.867 & 2.757 & 2.521 & 2.276 & 1.966 & 1.713 & 1.502 & 1.296 & 1.131 & 1.020 & 0.929 \\
GPT-2-medium & 3.148 & 2.891 & 2.874 & 2.735 & 2.663 & 2.547 & 2.369 & 2.122 & 1.886 & 1.645 & 1.424 & 1.225 & 1.068 & 0.943 & 0.855 \\
GPT-2-large & 2.937 & 2.888 & 2.740 & 2.764 & 2.589 & 2.515 & 2.362 & 2.128 & 1.837 & 1.618 & 1.401 & 1.254 & 1.094 & 0.948 & 0.887 \\
GPT-2-xl & 2.888 & 2.646 & 2.614 & 2.508 & 2.461 & 2.393 & 2.297 & 2.143 & 1.940 & 1.701 & 1.477 & 1.278 & 1.278 & 0.896 & 0.800 \\
LaMini-GPT-124M & 3.253 & 3.061 & 3.014 & 2.976 & 2.916 & 2.781 & 2.669 & 2.473 & 2.130 & 1.847 & 1.606 & 1.376 & 1.210 & 1.062 & 0.958 \\
LaMini-GPT-774M & 2.813 & 2.680 & 2.669 & 2.661 & 2.536 & 2.471 & 2.309 & 2.072 & 1.825 & 1.600 & 1.373 & 1.189 & 1.044 & 0.921 & 0.838 \\
LaMini-GPT-1.5B & 2.742 & 2.710 & 2.660 & 2.653 & 2.580 & 2.490 & 2.408 & 2.327 & 2.001 & 1.725 & 1.451 & 1.230 & 1.050 & 0.913 & 0.790 \\
Cerebras-GPT-111M & 3.348 & 3.034 & 2.943 & 2.878 & 2.796 & 2.716 & 2.607 & 2.455 & 2.249 & 2.012 & 1.792 & 1.595 & 1.393 & 1.170 & 0.957 \\
Cerebras-GPT-256M & 3.109 & 2.891 & 2.801 & 2.664 & 2.632 & 2.502 & 2.364 & 2.178 & 1.951 & 1.786 & 1.563 & 1.393 & 1.229 & 1.054 & 0.919 \\
Cerebras-GPT-1.3B & 2.610 & 2.789 & 2.628 & 2.521 & 2.388 & 2.315 & 2.238 & 2.097 & 1.926 & 1.732 & 1.595 & 1.459 & 1.316 & 1.156 & 1.030 \\
Cerebras-GPT-2.7B & 2.192 & 1.959 & 1.892 & 1.842 & 1.771 & 1.739 & 1.705 & 1.650 & 1.608 & 1.540 & 1.442 & 1.429 & 1.410 & 1.372 & 1.331 \\
Phi-1.5 & 2.641 & 2.883 & 2.652 & 2.428 & 2.361 & 2.152 & 1.961 & 1.802 & 1.634 & 1.468 & 1.317 & 1.201 & 1.088 & 0.981 & 0.901 \\
Phi-2 & 1.857 & 2.272 & 2.137 & 1.987 & 1.941 & 1.799 & 1.631 & 1.507 & 1.364 & 1.264 & 1.123 & 1.024 & 0.935 & 0.858 & 0.799 \\
OPT-350m & 3.199 & 3.117 & 2.972 & 2.972 & 2.784 & 2.621 & 2.438 & 2.157 & 1.890 & 1.637 & 1.426 & 1.271 & 1.119 & 1.004 & 0.881 \\
OPT-1.3b & 2.727 & 2.761 & 2.650 & 2.615 & 2.497 & 2.342 & 2.148 & 1.963 & 1.777 & 1.563 & 1.433 & 1.295 & 1.162 & 1.014 & 0.883 \\
OPT-2.7b & 2.495 & 2.480 & 2.441 & 2.391 & 2.331 & 2.277 & 2.106 & 1.987 & 1.817 & 1.652 & 1.530 & 1.391 & 1.289 & 1.188 & 1.081 \\
OPT-6.7b & 2.262 & 1.987 & 1.984 & 1.979 & 1.961 & 1.957 & 1.945 & 1.917 & 1.881 & 1.864 & 1.831 & 1.812 & 1.787 & 1.761 & 1.738 \\
ai-forever/mGPT & 2.285 & 2.089 & 2.086 & 2.093 & 2.071 & 2.043 & 2.018 & 2.007 & 1.996 & 1.941 & 1.919 & 1.867 & 1.833 & 1.786 & 1.753 \\
BART-base & 6.781 & 3.368 & 3.366 & 3.163 & 3.030 & 2.874 & 2.787 & 2.330 & 1.991 & 1.691 & 1.411 & 1.254 & 1.070 & 0.932 & 0.859 \\
BART-large & 4.145 & 3.214 & 3.202 & 3.056 & 2.953 & 2.689 & 2.490 & 2.121 & 1.796 & 1.524 & 1.296 & 1.105 & 0.957 & 0.828 & 0.758 \\
BART-large-cnn & 6.028 & 3.223 & 3.103 & 3.029 & 2.829 & 2.602 & 2.285 & 1.963 & 1.739 & 1.485 & 1.270 & 1.104 & 0.962 & 0.858 & 0.771 \\
BART-large-xsum & 4.263 & 3.161 & 3.093 & 2.973 & 2.847 & 2.643 & 2.371 & 2.092 & 1.806 & 1.510 & 1.310 & 1.129 & 0.980 & 0.857 & 0.774 \\
T5-small & 4.384 & 1.251 & 1.223 & 1.135 & 1.048 & 0.991 & 0.958 & 0.903 & 0.845 & 0.803 & 0.781 & 0.749 & 0.717 & 0.664 & 0.641 \\
T5-base & 4.798 & 1.174 & 1.060 & 1.037 & 0.950 & 0.885 & 0.835 & 0.776 & 0.745 & 0.734 & 0.684 & 0.644 & 0.626 & 0.591 & 0.575 \\
mT5-base & 16.143 & 2.879 & 2.822 & 2.781 & 2.722 & 2.692 & 2.671 & 2.578 & 2.471 & 2.451 & 2.388 & 2.322 & 2.245 & 2.162 & 2.079 \\
mT5-large & 21.711 & 2.841 & 2.814 & 2.776 & 2.711 & 2.687 & 2.648 & 2.560 & 2.472 & 2.412 & 2.290 & 2.211 & 2.129 & 2.032 & 1.941 \\
T5-v1.1-base & 10.500 & 1.389 & 1.261 & 1.225 & 1.176 & 1.123 & 1.053 & 0.991 & 0.930 & 0.868 & 0.808 & 0.743 & 0.680 & 0.622 & 0.561 \\
switch-base-8 & 27.451 & 1.561 & 1.472 & 1.374 & 1.251 & 1.223 & 1.125 & 1.050 & 0.981 & 0.923 & 0.849 & 0.791 & 0.741 & 0.689 & 0.651 \\
switch-base-16 & 21.009 & 1.389 & 1.290 & 1.203 & 1.187 & 1.094 & 1.044 & 0.991 & 0.913 & 0.866 & 0.807 & 0.756 & 0.745 & 0.666 & 0.631 \\
switch-base-32 & 18.065 & 1.351 & 1.262 & 1.172 & 1.112 & 1.042 & 0.962 & 0.901 & 0.847 & 0.788 & 0.733 & 0.681 & 0.642 & 0.601 & 0.567 \\
\bottomrule
\end{tabular}
}
\end{table*}

\begin{figure}[!hbp]
    \centering
    \includegraphics[width=0.7\linewidth]{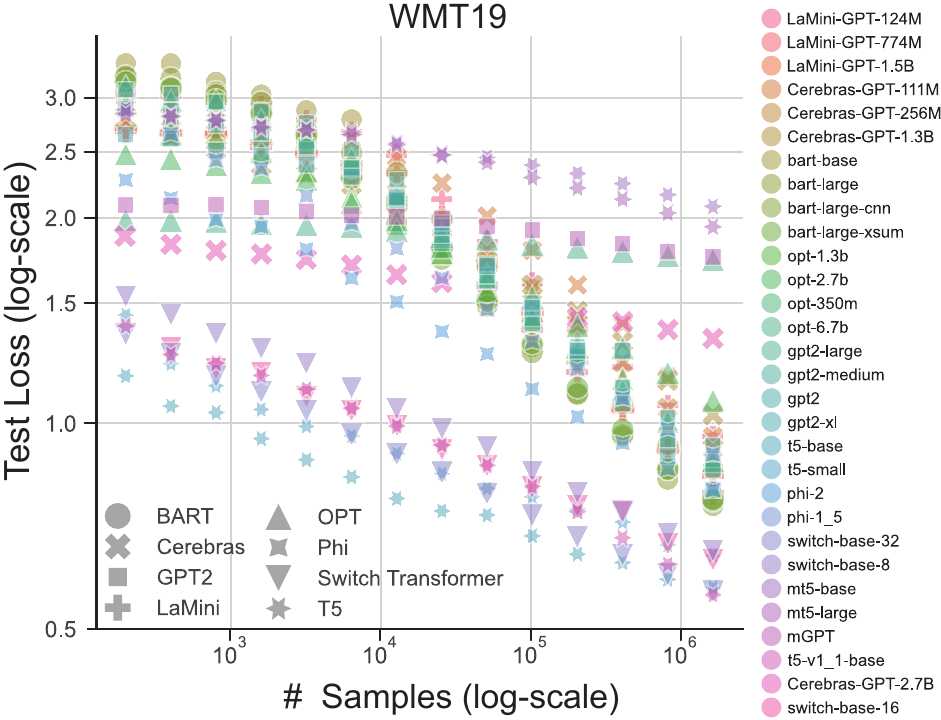}
    \caption{The test losses of $30$ models fine-tuned on various sizes of subsets derived from the WMT19 dataset. The point size reflects the corresponding model size.}
    \label{fig:wmt}
\end{figure}

\newpage

\begin{table*}[htb]
    \centering
    \small
    \captionsetup{justification=centering}
    \caption{Test loss of $30$ models fine-tuned on subsets of Gigaword dataset. The data size ranges from $0$ to $1638400$.}
    \resizebox{\linewidth}{!}{
\begin{tabular}{lccccccccccccccc}
\toprule
\multicolumn{1}{c}{Model} & ~~~~0~~~~ & ~~~200~~~ & ~~~400~~~ & ~~~800~~~ & ~~~1600~~~ & ~~~3200~~~ & ~~~6400~~~ & ~~12800~~ & ~~25600~~ & ~~51200~~ & 102400 & 204800 & 409600 & 819200 & 1638400 \\
\midrule
GPT-2 & 4.147 & 2.691 & 2.596 & 2.516 & 2.429 & 2.329 & 2.204 & 2.099 & 1.983 & 1.883 & 1.777 & 1.690 & 1.597 & 1.508 & 1.431 \\
GPT-2-medium & 3.723 & 2.298 & 2.214 & 2.130 & 2.050 & 1.965 & 1.891 & 1.810 & 1.742 & 1.672 & 1.602 & 1.530 & 1.465 & 1.398 & 1.349 \\
GPT-2-large & 3.613 & 2.154 & 2.103 & 2.018 & 1.961 & 1.887 & 1.799 & 1.750 & 1.671 & 1.603 & 1.540 & 1.479 & 1.408 & 1.354 & 1.305 \\
GPT-2-xl & 3.411 & 2.044 & 2.010 & 1.954 & 1.880 & 1.814 & 1.773 & 1.702 & 1.634 & 1.577 & 1.521 & 1.468 & 1.413 & 1.356 & 1.286 \\
LaMini-GPT-124M & 4.414 & 2.645 & 2.546 & 2.457 & 2.384 & 2.300 & 2.203 & 2.110 & 1.996 & 1.888 & 1.790 & 1.694 & 1.595 & 1.511 & 1.438 \\
LaMini-GPT-774M & 4.161 & 2.142 & 2.085 & 2.015 & 1.942 & 1.873 & 1.814 & 1.746 & 1.673 & 1.603 & 1.541 & 1.480 & 1.422 & 1.358 & 1.308 \\
LaMini-GPT-1.5B & 4.053 & 2.041 & 2.000 & 1.927 & 1.877 & 1.824 & 1.766 & 1.703 & 1.645 & 1.570 & 1.518 & 1.459 & 1.439 & 1.354 & 1.299 \\
Cerebras-GPT-111M & 5.108 & 3.505 & 3.362 & 3.217 & 3.080 & 2.939 & 2.780 & 2.658 & 2.507 & 2.354 & 2.208 & 2.048 & 1.914 & 1.796 & 1.677 \\
Cerebras-GPT-256M & 4.574 & 3.043 & 2.934 & 2.823 & 2.686 & 2.576 & 2.473 & 2.350 & 2.225 & 2.112 & 1.994 & 1.888 & 1.785 & 1.683 & 1.586 \\
Cerebras-GPT-1.3B & 3.834 & 2.401 & 2.324 & 2.257 & 2.193 & 2.139 & 2.082 & 2.008 & 1.924 & 1.851 & 1.770 & 1.682 & 1.618 & 1.550 & 1.482 \\
Cerebras-GPT-2.7B & 3.400 & 2.125 & 2.054 & 1.983 & 1.933 & 1.866 & 1.806 & 1.745 & 1.692 & 1.637 & 1.576 & 1.533 & 1.480 & 1.440 & 1.391 \\
Phi-1.5 & 4.169 & 2.354 & 2.266 & 2.157 & 2.069 & 1.992 & 1.905 & 1.834 & 1.761 & 1.679 & 1.607 & 1.540 & 1.483 & 1.410 & 1.361 \\
Phi-2 & 3.245 & 1.788 & 1.747 & 1.705 & 1.674 & 1.639 & 1.602 & 1.574 & 1.534 & 1.478 & 1.453 & 1.431 & 1.389 & 1.354 & 1.319 \\
OPT-350m & 3.848 & 2.422 & 2.312 & 2.227 & 2.149 & 2.078 & 2.013 & 1.928 & 1.858 & 1.768 & 1.712 & 1.635 & 1.574 & 1.512 & 1.450 \\
OPT-1.3b & 3.163 & 1.879 & 1.828 & 1.772 & 1.722 & 1.686 & 1.638 & 1.588 & 1.543 & 1.491 & 1.446 & 1.403 & 1.368 & 1.327 & 1.290 \\
OPT-2.7b & 2.971 & 1.734 & 1.697 & 1.658 & 1.620 & 1.576 & 1.541 & 1.502 & 1.462 & 1.429 & 1.391 & 1.363 & 1.330 & 1.301 & 1.270 \\
OPT-6.7b & 2.862 & 1.694 & 1.656 & 1.623 & 1.582 & 1.549 & 1.506 & 1.460 & 1.428 & 1.400 & 1.368 & 1.339 & 1.308 & 1.276 & 1.245 \\
ai-forever\/mGPT & 3.676 & 2.379 & 2.386 & 2.238 & 2.186 & 2.034 & 1.939 & 1.863 & 1.802 & 1.732 & 1.651 & 1.586 & 1.530 & 1.452 & 1.379 \\
BART-base & 8.663 & 3.299 & 3.120 & 2.884 & 2.710 & 2.535 & 2.391 & 2.021 & 1.894 & 1.797 & 1.696 & 1.630 & 1.548 & 1.469 & 1.408 \\
BART-large & 4.727 & 2.211 & 2.102 & 1.984 & 1.895 & 1.809 & 1.734 & 1.666 & 1.610 & 1.537 & 1.483 & 1.420 & 1.361 & 1.303 & 1.257 \\
BART-large-CNN & 4.619 & 2.268 & 2.172 & 2.063 & 1.949 & 1.842 & 1.737 & 1.670 & 1.594 & 1.524 & 1.472 & 1.403 & 1.364 & 1.306 & 1.255 \\
BART-large-XSUM & 4.486 & 2.204 & 2.128 & 2.030 & 1.934 & 1.839 & 1.751 & 1.686 & 1.613 & 1.546 & 1.484 & 1.412 & 1.371 & 1.311 & 1.261 \\
T5-small & 3.675 & 2.078 & 2.061 & 2.028 & 1.911 & 1.863 & 1.804 & 1.743 & 1.680 & 1.624 & 1.554 & 1.484 & 1.406 & 1.322 & 1.250 \\
T5-base & 2.880 & 1.758 & 1.725 & 1.679 & 1.638 & 1.597 & 1.542 & 1.492 & 1.444 & 1.395 & 1.351 & 1.301 & 1.247 & 1.196 & 1.146 \\
mT5-base & 11.509 & 2.810 & 2.689 & 2.589 & 2.432 & 2.292 & 2.167 & 2.024 & 1.851 & 1.721 & 1.599 & 1.482 & 1.371 & 1.253 & 1.148 \\
mT5-large & 10.154 & 2.567 & 2.462 & 2.331 & 2.212 & 2.110 & 1.987 & 1.890 & 1.781 & 1.679 & 1.588 & 1.492 & 1.418 & 1.332 & 1.259 \\
T5-v1.1-base & 9.205 & 2.582 & 2.451 & 2.283 & 2.123 & 1.979 & 1.870 & 1.717 & 1.614 & 1.502 & 1.414 & 1.326 & 1.241 & 1.151 & 1.071 \\
switch-base-8 & 20.602 & 2.672 & 2.573 & 2.286 & 2.124 & 1.991 & 1.859 & 1.726 & 1.619 & 1.512 & 1.430 & 1.356 & 1.275 & 1.206 & 1.149 \\
switch-base-16 & 17.835 & 2.641 & 2.443 & 2.253 & 2.035 & 1.916 & 1.789 & 1.675 & 1.583 & 1.480 & 1.395 & 1.334 & 1.260 & 1.196 & 1.123 \\
switch-base-32 & 14.677 & 2.430 & 2.309 & 2.187 & 1.967 & 1.881 & 1.734 & 1.625 & 1.563 & 1.457 & 1.383 & 1.305 & 1.246 & 1.186 & 1.106 \\
\bottomrule
\end{tabular}
}
\end{table*}

\begin{figure}[!hbp]
    \centering
    \includegraphics[width=0.7\linewidth]{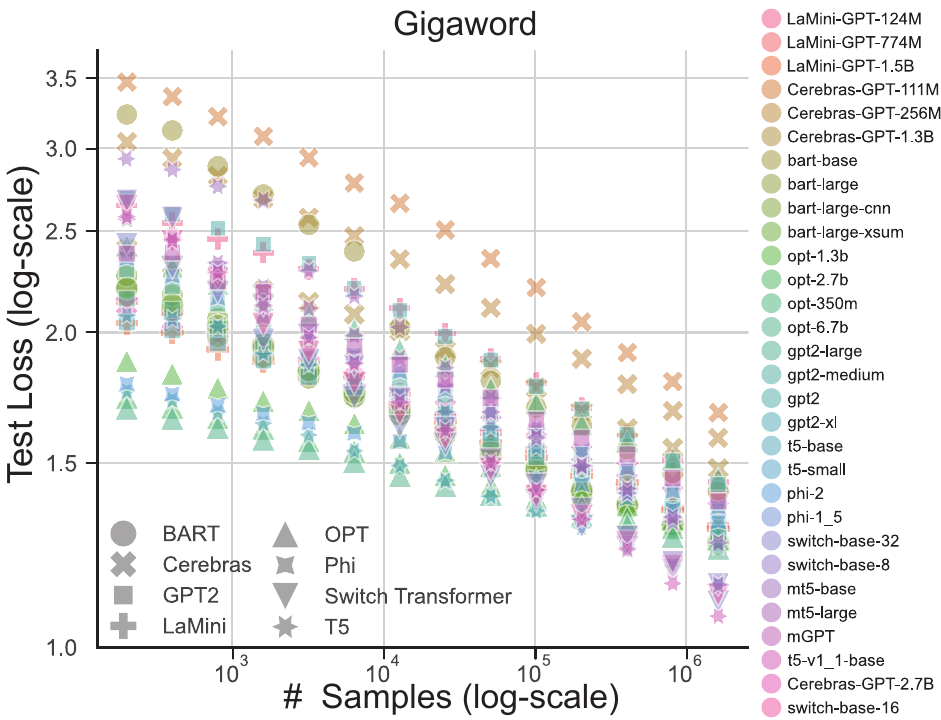}
    \caption{The test losses of $30$ models fine-tuned on various sizes of subsets derived from the Gigaword dataset. The point size reflects the corresponding model size.}
    \label{fig:giga}
\end{figure}
\newpage

\section{Full Analysis Studies}
\subsection{Influence of Hyper-parameters}
\label{app:ablation_hyper}

\begin{figure}[!hbp]
    \centering
    \includegraphics[width=\linewidth]{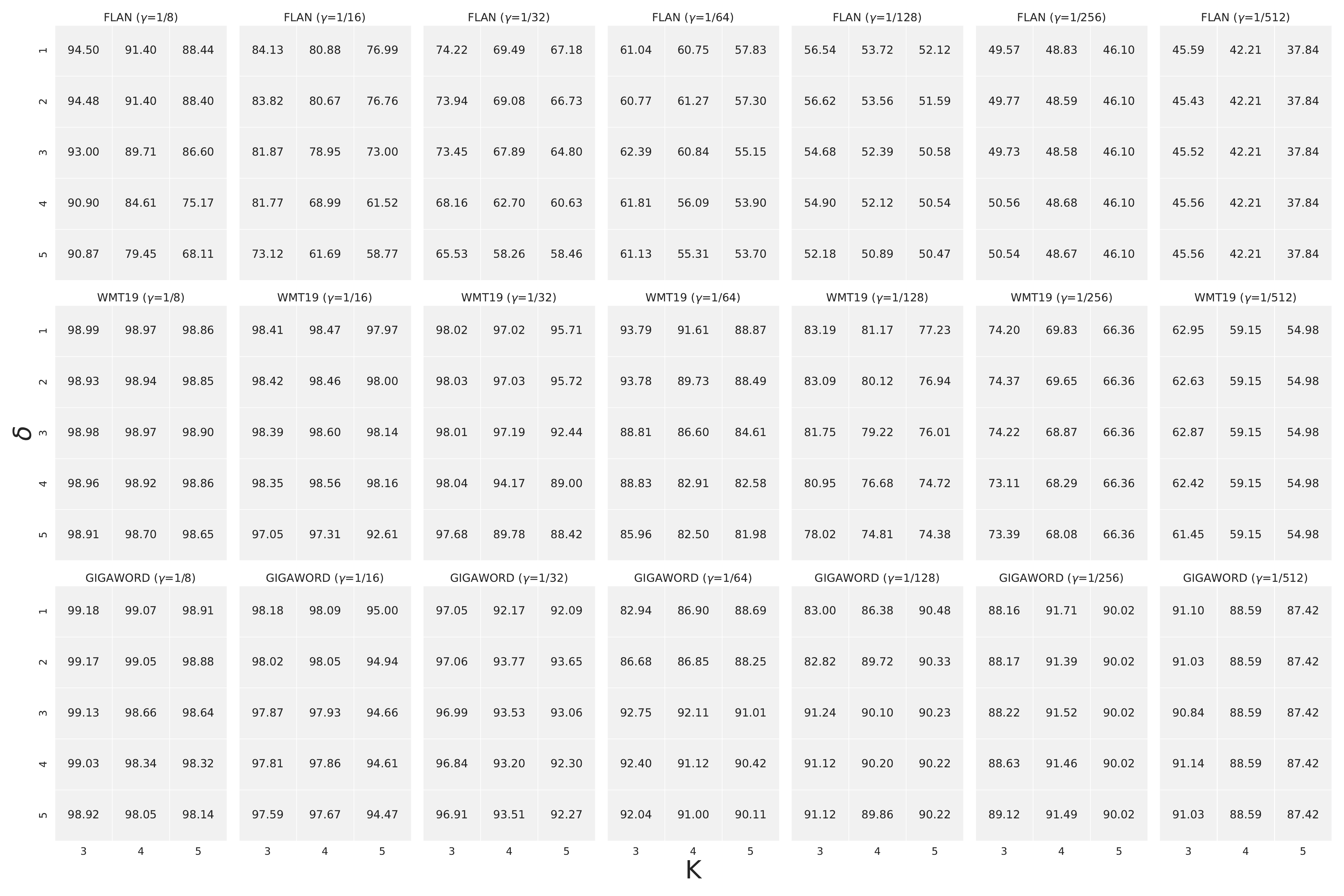}
    \caption{\textbf{PearCorr} of \emph{AtS} with varied hyper-parameters $\delta$ and $k$ across FLAN, WMT19 and Gigaword datasets. Each block presents an ablation analysis, delineating the impact of hyper-parameter settings on specific subsets.}
    \label{fig:ablationstudyhyperfull}
\end{figure}

\begin{table}[h]
    \centering
    \begin{tabular}{c|ccccc}
\toprule
GPT-2 &	$4.386\pm 0.0016$ & $4.288\pm 0.0018$ & $4.191\pm 0.0015$ & $4.060\pm 0.0011$ & $3.890\pm 0.0013$\\
Cerebras-256M & $3.393\pm 0.0021$ & $3.319 \pm 0.0022$ & $3.230\pm 0.0012$ & $3.127\pm 0.0010$ & $3.054\pm 0.0009$\\
BART-base & $4.159\pm 0.0051$ & $3.990\pm 0.0049$ & $3.850\pm 0.0045$ & $3.685 \pm 0.0042$ & $3.532\pm 0.0020$\\
OPT-350M & $3.203\pm 0.0025$ & $3.132\pm 0.0023$& $3.020\pm 0.0021$ & $2.943\pm 0.0016$& $2.848\pm 0.0013$\\
\bottomrule
    \end{tabular}
    \caption{Variance of fine tuning results of four typical models on FLAN. It is shown that the fine tuning processes are very stable.}
    \label{tab:variance}
\end{table}

\newpage
\subsection{\emph{AtS} on Stratified $\gM$}
\begin{figure}[!hb]
    \centering
    
    \includegraphics[width=0.9\linewidth]{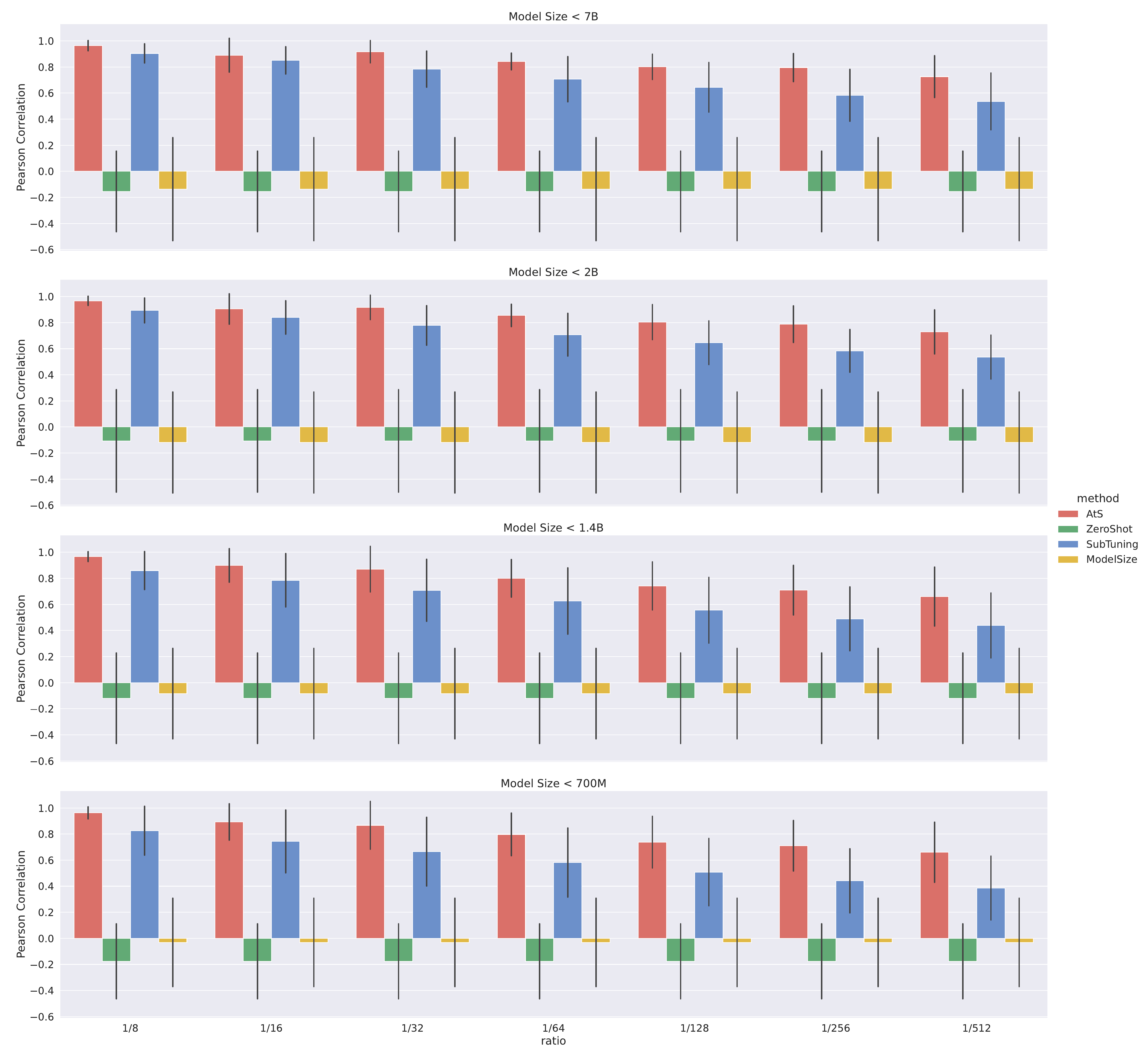}
    \caption{Performance of \emph{AtS} on stratified $\gM$ with varied memory budgets measured by \textbf{PearCorr}. }
    \label{fig:app_ablation_size}
\end{figure}

Here, we present comprehensive results demonstrating the efficacy of \emph{AtS} on stratified $\mathcal{M}$ across four distinct memory budgets: $7B$, $2B$, $1.4B$, and $700M$, as depicted in Figure \ref{fig:app_ablation_size}. Each of these memory budgets corresponds to different subsets of the $\gM$, comprising $30$, $25$, $21$, and $15$ individual models, respectively. Notably, \emph{AtS} consistently demonstrates superior performance across all memory budgets, affirming its practical viability for real-world deployment scenarios.

\label{app:ablation_size}       

\newpage
\subsection{LLM Selection by Fitting Scaling Law}
\label{app:ablation_different_laws}
Here, we present the full selection results of three scaling-law-based selection methods on three datasets in Table \ref{app:selection-by-scaling-law}. Examining both metrics, we observe that the \emph{AtS} method consistently outperforms the other two methods (\emph{OurFit} and \emph{VanillaFit}) across all datasets and budget ratios. It again demonstrates the robustness and stability of our proposed method.

\begin{table}[htb]
    \caption{Model selection results (\textbf{PearCorr}, \textbf{RelAcc}) of three scaling-law-based methods on three datasets (FLAN, WMT19, Gigaword) in percentage. The best result within the same dataset and budget ratio is in \textbf{bold} font, and the second best result is \underline{underlined}.}
    \label{app:selection-by-scaling-law}
    \small
    \centering
    \resizebox{\linewidth}{!}{
    \begin{tabular}{c|c|ccc|ccc|ccc}
\hline
\multicolumn{1}{c}{~} & \multicolumn{1}{c|}{~} & \multicolumn{3}{c|}{\textbf{FLAN}} & \multicolumn{3}{c|}{\textbf{WMT19}} & \multicolumn{3}{c}{\textbf{Gigaword}} \\
Metric & ~~Ratio~~ & ~~~~\emph{AtS}~~~~ & ~~\emph{OurFit}~~ & \emph{VanillaFit} & ~~~~\emph{AtS}~~~~ & ~~\emph{OurFit}~~ & \emph{VanillaFit} & ~~~~\emph{AtS}~~~~ & ~~\emph{OurFit}~~ & \emph{VanillaFit} \\
\hline
\multirow{8}{*}{\textbf{PearCorr} (\%)} & 1/8 & \textbf{90.9} & \underline{77.9} & 34.7 & \textbf{98.9} & \underline{95.0} & 94.4 & \textbf{98.9} & \underline{97.0} & 95.1 \\
~ & 1/16 & \textbf{73.1} & \underline{67.4} & 58.1 & \textbf{97.0} & \underline{93.6} & 83.7 & \textbf{97.6} & 90.7 & \underline{92.8} \\
~ & 1/32 & \textbf{65.5} & \underline{54.4} & 43.1 & \textbf{97.7} & \underline{91.1} & 79.6 & \textbf{96.9} & 88.3 & \underline{91.0} \\
~ & 1/64 & \textbf{61.1} & \underline{47.6} & 46.7 & \textbf{86.0} & \underline{83.9} & 30.9 & \textbf{92.0} & 83.6 & \underline{84.3} \\
~ & 1/128 & \underline{52.2} & \textbf{54.9} & 41.4 & \underline{78.0} & \textbf{78.9} & 35.2 & \textbf{91.1} & \underline{83.6} & 47.3 \\
~ & 1/256 & \textbf{50.5} & 41.1 & \underline{45.0} & \textbf{73.4} & \underline{72.9} & 41.1 & \textbf{89.1} & 81.5 & \underline{85.8} \\
~ & 1/512 & \textbf{45.6} & \underline{36.8} & 20.7 & \underline{61.5} & \textbf{61.5} & 56.5 & \textbf{91.0} & 78.5 & \underline{79.3} \\

\hline

\multirow{8}{*}{\textbf{RelAcc} (\%)} & 1/8 & \underline{93.6} & \textbf{100.0} & 39.0 & \underline{99.1} & 84.9 & \textbf{99.6} & \textbf{100.0} & \textbf{100.0} & \textbf{100.0} \\
~ & 1/16 & \underline{93.2} & \textbf{100.0} & \underline{93.2} & \textbf{99.1} & \underline{84.9} & 80.7 & 91.4 & \textbf{100.0} & \textbf{100.0} \\
~ & 1/32 & \underline{93.2} & \textbf{100.0} & \underline{93.2} & \textbf{99.6} & 78.5 & \textbf{99.6} & 94.3 & \textbf{100.0} & \textbf{100.0} \\
~ & 1/64 & \underline{93.2} & \textbf{100.0} & 90.7 & \textbf{99.1} & 81.8 & \textbf{99.1} & \textbf{100.0} & 94.3 & \textbf{100.0} \\
~ & 1/128 & \underline{85.3} & \underline{85.3} & \textbf{93.2} & \textbf{99.1} & 78.5 & \textbf{99.1} & \textbf{94.3} & \textbf{94.3} & \textbf{94.3} \\
~ & 1/256 & \textbf{93.2} & \underline{85.3} & \underline{85.3} & \textbf{99.1} & 77.6 & \textbf{99.1} & \textbf{94.3} & 87.2 & \textbf{94.3} \\
~ & 1/512 & \textbf{93.2} & 85.3 & \textbf{93.2} & \textbf{99.1} & 77.6 & \textbf{99.1} & \textbf{91.4} & \textbf{91.4} & 87.3 \\
\bottomrule
\end{tabular}
}

\end{table}

\newpage

\subsection{\textit{AtS-Family}: a Variant with Model Family Prior}

While the main idea of \textit{AtS}  is to select LLMs based on the Scaling Law, it can also be integrated with other methods or heuristics to simplify the selection process and significantly reduce computational costs. We introduce a variant, \textit{AtS-Family}, which combines \textit{AtS} with the intuitive hypothesis that \emph{larger models within a family tend to exhibit superior performance}. Specifically, \textit{AtS-Family} narrows the candidate model set to the largest model within each model family (e.g. GPT-2-xl in the GPT-2 family, OPT-6.7b in the OPT family) and subsequently applies \textit{AtS} to the limited candidates. This approach markedly reduces computational complexity, as it necessitates fine-tuning only a single model per family. The efficacy of \textit{AtS-Family} is demonstrated in Table \ref{tab:Ats-Family}.

\begin{table}[htbp]
    \centering
\small
    \caption{Model selection results of \textit{AtS-Family} evaluated by \textbf{RelAcc} on three datasets(FLAN, WMT19, Gigaword) in percentage. The best result within the same dataset and budget ratio is in \textbf{bold} font, and the second best result is \underline{underlined}.}

\begin{tabular}{c|ccc|ccc|ccc}
\hline
\multicolumn{1}{c|}{~} & \multicolumn{3}{c|}{\textbf{FLAN}} & \multicolumn{3}{c|}{\textbf{WMT19}} & \multicolumn{3}{c}{\textbf{Gigaword}} \\
Ratio& ~~~~\textit{AtS}~~~~ & \textit{SubTuning} & \textit{AtS-Family} & ~~~~\textit{AtS}~~~~ & \textit{SubTuning} & \textit{AtS-Family} & ~~~~\textit{AtS}~~~~ & \textit{SubTuning} & \textit{AtS-Family} \\
\hline
 1/8 & \textbf{93.6} & \underline{93.2} & \underline{93.2} & \underline{99.1} & \underline{99.1} & \textbf{99.6} & \textbf{100.0} & 87.6 & \underline{94.3} \\
 1/16 & \textbf{93.2} & \textbf{93.2} & \textbf{93.2} & \underline{99.1} & \underline{99.1} & \textbf{99.6} & \underline{91.4} & 87.6 & \textbf{94.3} \\
 1/32 & \textbf{93.2} & \textbf{93.2} & \textbf{93.2} & \textbf{99.6} & 99.1 & \textbf{99.6} & \textbf{94.3} & 87.6 & \textbf{94.3} \\
 1/64 & \textbf{93.2} & \textbf{93.2} & \textbf{93.2} & \underline{99.1} & \underline{99.1} & \textbf{99.6} & \textbf{100.0} & 71.3 & \underline{94.3} \\
 1/128 & \underline{85.3} & 59.6 & \textbf{93.2} & \underline{99.1} & \underline{99.1} & \textbf{99.6} & \textbf{94.3} & 71.3 & \textbf{94.3} \\
 1/256 & \textbf{93.2} & 59.6 & \textbf{93.2} & \underline{99.1} & \underline{99.1} & \textbf{99.6} & \textbf{94.3} & 71.3 & \textbf{94.3} \\
 1/512 & \textbf{93.2} & 59.6 & \textbf{93.2} & \underline{99.1} & \underline{99.1} & \textbf{99.6} & \underline{91.4} & 71.3 & \textbf{94.3} \\
\hline
\end{tabular}
    \label{tab:Ats-Family}
\end{table}

\end{document}